# Machine Learning with Knowledge Constraints for Process Optimization of Open-Air Perovskite Solar Cell Manufacturing

Zhe Liu[1,*,§], Nicholas Rolston[2,*,], Austin C. Flick[2], Thomas W. Colburn[2], Zekun Ren[3],

Reinhold H. Dauskardt[2,†], Tonio Buonassisi[1,†]

[1]Massachusetts Institute of Technology, Cambridge, MA, United States
[2]Stanford University, Stanford, CA, United States
[3]Singapore-MIT Alliance for Research and Technology, Singapore

*These authors contributed equally as co-first authors.
†Email correspondence: buonassi@mit.edu (T.B.) & rhd@stanford.edu (R.H.D)
§This author is now at Northwestern Polytechnical University (NPU), Xi'an, Shaanxi, P.R. China

**Abstract**

Perovskite photovoltaics (PV) have achieved rapid development in the past decade in terms of power conversion efficiency of small-area lab-scale devices; however, successful commercialization still requires further development of low-cost, scalable, and high-throughput manufacturing techniques. One of the critical challenges of developing a new fabrication technique is the high-dimensional parameter space for optimization, but machine learning (ML) can readily be used to accelerate perovskite PV scaling. Herein, we present an ML-guided framework of sequential learning for manufacturing process optimization. We apply our methodology to the Rapid Spray Plasma Processing (RSPP) technique for perovskite thin films in ambient conditions. With a limited experimental budget of screening 100 process conditions, we demonstrated an efficiency improvement to 18.5% as the best-in-our-lab device fabricated by RSPP, and we also experimentally found 10 unique process conditions to produce the top-performing devices of more than 17% efficiency, which is 5 times higher rate of success than the control experiments with pseudo-random Latin hypercube sampling. Our model is enabled by three innovations: (a) flexible knowledge transfer between experimental processes by incorporating data from prior experimental data as a probabilistic constraint; (b) incorporation of both subjective human observations and ML insights when selecting next experiments; (c) adaptive strategy of locating the region of interest using Bayesian optimization first, and then conducting local exploration for high-efficiency devices. Furthermore, in virtual benchmarking, our framework achieves faster improvements with limited experimental budgets than traditional design-of-experiments methods (*e.g.*, one-variable-at-a-time sampling). This framework shows the capability of incorporating researchers' domain knowledge into the ML-guided optimization loop; therefore, it has the potential to facilitate the wider adoption of ML in scaling to perovskite PV manufacturing.

## 1. Introduction:

Metal halide perovskites are efficient solar absorbers compatible with low-cost solution processing methods that show promise as an emerging thin-film photovoltaic (PV) technology. Scaling up the fabrication process is currently one of the critical research areas of perovskite technology on the path to commercialization [1]. Despite the success of >25% efficient perovskite solar cells in academic labs using spin coating [2,3], this processing method is not scalable to a manufacturing line. Recently,

Rolston et al. [4] demonstrated a high-throughput rapid spray-plasma processing (RSPP) method as a scalable open-air fabrication process because it has the potential of achieving low-cost perovskite PV modules at a manufacturing cost of ~$0.2 /W. In addition to cost, the main advantage of this spray-deposition-based technique is its ultra-high throughput and improved mechanical properties of the thin film in comparison to other scalable processing methods, such as blade coating, slot-die coating, and roll-to-roll printing.

For new scalable perovskite manufacturing processes (including RSPP), it typically takes months to years to achieve process control and reproducibility on the module-scale, and several years to estimate the upper potential of the technology. One of the key challenges is that there are many processing parameters to co-optimize [1] — e.g., precursor composition, speed, temperature, head/nozzle height, curing methods. This high-dimensional optimization problem can be nearly impossible to solve by brute force, or even by the sophisticated design of experiments.

Sequential machine learning methods, e.g., Bayesian optimization (BO), have emerged as effective optimization strategies to explore a wide range of chemical reaction synthesis [5–7] and material optimization [8–16]. BO has been shown to work well for optimization problems with under 20 variables [17] or up to 30 variables with some algorithm modifications [18,19]. Therefore, we have chosen to study sequential-learning-based optimization strategies in our current study of RSPP perovskite PV devices.

Current reports of sequential learning studies with classical Bayesian optimization have two common drawbacks: (1) no direct channel to incorporate information from previous relevant studies, and (2) inflexibility to adapt researchers' qualitative feedbacks into the iterative loop. On the one hand, the ML model sometimes requires a significant amount of data to learn what has already become apparent to the researchers [20]. On the other hand, these drawbacks could discourage materials science researchers to adopt machine learning tools because their domain knowledge could not be utilized in the classical BO iterations.

Previous data and researchers' knowledge are useful information sources for experimental planning. An acquisition function in a BO framework is the "decision-maker" to produce an experimental plan in the next round. Therefore, we can incorporate domain knowledge as probabilistic constraints for the acquisition function (introduced by Gelbart et al. [21]). With the aim of rapid optimization in materials science, Sun and Tiihonen et al. incorporated density functional theory (DFT) calculations of phase stability as a probabilistic constraint for the experimental acceleration of composition optimization to improve perovskite stability, avoiding compositions susceptible to phase segregation [22].

Simple information such as a visual assessment of film thickness, color, and structural defects can be a powerful tool that provides additional guidance for intelligent and efficient optimization. In other words, if an experienced researcher identifies a low-quality perovskite film, the subsequent device fabrication is no longer necessary. Defining a probabilistic constraint is an effective and flexible way to incorporate this information into a sequential learning framework.

In this work, we develop a sequential learning framework with probabilistic constraints for rapid process optimization with power conversion efficiency (PCE) as the target variable. As illustrated in Fig. 1, the sequential learning framework iteratively learns the process-efficiency relation and suggests new experiments to achieve the optimal target. For a general framework, we start with experimental planning of process conditions with a model-free sampling method for the initial round. Then, perovskite solar cells are fabricated by the RSPP method and PCE is measured with a solar simulator under standard testing conditions. With the experimental data of the process parameters and the device PCEs, we train the regression model to learn the process-efficiency relation; and the regression model is subsequently used to predict the PCE (and its prediction uncertainty) for the unsampled regions. Finally, the prediction results are evaluated by an acquisition function together with the constraint information, and therefore a new round of experiments is planned. By including two additional knowledge constraints (*i.e.*, visual assessment of film quality and previous experimental data from a related study), the optimization framework maps out the parameter space and tends to avoid the less promising regions based on observational film-quality information and prior device data. Hence, sampling suggestions focus on the most promising regions within the parameter space.

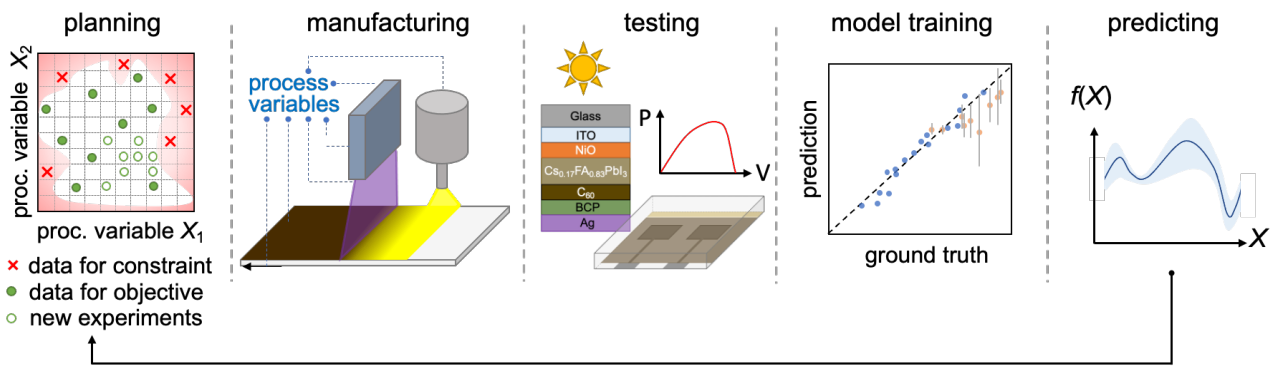

**Figure 1. Schematic of sequential learning optimization of perovskite solar cells with probabilistic constraints.** The five-step workflow is planning, manufacturing, testing, model training and prediction. This workflow iterates until the target efficiency is achieved or the maximum experimental budget is reached.

To demonstrate the capability of our optimization approach, we consider six key RSPP input variables for the perovskite absorber layer in a device fabrication process. We aimed to exceed the best PCE of solar cells produced with our RSPP. First, we show that efficiency improvements can be obtained with our sequential-learning framework, reaching 18.5% efficiency in five experimental iterations. Second, we describe how multiple sources of information were fused into our sequential learning framework as a probabilistic constraint. Third, we analyze the learned relationships between input variables and efficiency, extracting some generalizable insights. Fourth, we benchmark the acceleration factor and enhancement factor of our optimization process against conventional model-free sampling methods for the design of experiments sampling methods with optimization simulations, demonstrating an excellent acceleration within the limited experimental budget of fewer than 100 conditions.

## 2. Methods:

2.1 Experiments

The details of solar cell fabrication with an RSPP tool have been discussed elsewhere [4,23,24]. Here is a brief overview of the fabrication process. A NiO$_x$ layer with a thickness of ~20 nm was deposited by spin coating onto an ITO-coated glass substrate (Xin Yan Technologies, 10 Ω/□). The precursor used for NiO$_x$ deposition was Ni(NO$_3$)$_2$ in 94% vol. ethylene glycol with 6% vol. of ethylenediamine. A perovskite absorber layer of Cs$_{0.17}$FA$_{0.83}$PbI$_3$ was then deposited with RSPP using an impact ultrasonic spray nozzle (Sonotek) connected in-line to a slot-type atmospheric plasma jet (Plasmatreat). The basic working principle is to uniformly deposit the precursor solution with the spray nozzle onto a heated glass substrate followed by brief exposure to the afterglow of a nitrogen plasma to achieve rapid solidification and form a polycrystalline thin film. The precursors of CsI (Sigma-Aldrich), FAI (Sigma-Aldrich), and PbI$_2$ (TCI) were dissolved in the mixed solvent of anhydrous N,N-dimethylformamide (DMF, Acros), and dimethyl sulfoxide (DMSO, Acros) with a volume ratio of 1:2. The molar concentration of the precursors is 0.15 M. Subsequently, a fullerene (C$_{60}$, MER Corp) layer with a thickness of 45 nm, Bathocuproine (BCP, Sigma-Aldrich) layer with a thickness of 7.5 nm, and silver (Ag) electrode with a thickness of 150 nm was deposited by thermal evaporation (Angstrom Amod evaporator) in sequence. The area of each perovskite solar cell is 0.21 cm$^2$, which is defined by the stencil mask during Ag deposition. This process flow results in a p-i-n solar cell structure, with seven-layer stacks illuminated through the substrate: glass/ ITO/ NiO$_x$/ Cs$_{0.17}$FA$_{0.83}$PbI$_3$/ C$_{60}$/ BCP/ Ag. The current-voltage (*I-V*) characteristics of the solar cell devices were characterized under the standard testing condition (STC), *i.e.*, at 25 °C under the illumination of an AM1.5G solar spectrum. The *I-V* curves for each device were measured from 1.2V to -0.1V (a reverse scan) at 0.05 V intervals with a 0.1 s delay at each voltage step. For measurement consistency, there was no preconditioning (*e.g.*, light illumination, voltage bias) before the *I-V* scans were taken.

2.2 Machine Learning

The BO method was used for process optimization. As inputs, we selected six RSPP process variables that affect perovskite absorber layer deposition: substrate temperature (°C), linear speed of the spray and plasma nozzles (cm/s), spray flow rate of precursor liquid (mL/min), gas flow rate into plasma nozzle (L/min), height of plasma nozzle (cm), and plasma duty cycle (%, corresponding to the ratio of time that the plasma receives DC power). The target variable (*i.e.*, output) was solar cell power conversion efficiency (PCE). The optimization bounds were chosen as shown in Table I. Given the sampling intervals, a full-grid sampling can result in 41,580 unique process conditions.

These six process variables and their optimization bounds were chosen based on historical data of more than 300 process conditions (in a similar setup with different spray and plasma nozzles). The less explored variable in the past was set to a large range, *e.g.*, plasma duty cycle. We have given careful expert consideration when defining this parameter space for optimization, intending to balance experimental feasibility and broadness of the parameter space. The researcher-defined parameter space typically works very well in Bayesian-optimization-guided experimental work [15].

Table I: Optimization ranges for the six process variables (model inputs)

| Process Variable | Total Range (Interval) | Process Variable | Total Range (Interval) |
|---|---|---|---|
| Temperature | 125 – 175 °C (5°C) | Plasma height | 0.8 – 1.2 cm (0.2 cm) |
| Speed | 10 – 30 cm/s (2.5 cm/s) | Plasma gas flow | 15 – 35 L/min (5 L/min) |
| Spray flow | 2.0 – 5.0 mL/s (0.5 mL/min) | Plasma duty cycle | 25 – 100 % (25%) |

Sequential learning for process optimization has five steps: initial sampling, regression, prediction, evaluation, and acquisition. The initial round was done by Latin hypercube sampling (LHS), which is a pseudo-random method that samples the parameter space relatively uniformly in high dimensions. The surrogate model was Gaussian Process (GP) regression with anisotropic Matern52. The choice of GP affords slightly better extrapolation than step-wise regressors like Random Forest regression [25]; the anisotropic kernel allows the hyperparameters of each input variable to be independently tuned, so that some variables can have a stronger impact on the predicted output variable (PCE) than others [26].

The acquisition function consists of two parts. The first part is the objective function (*i.e.*, process-efficiency relation). We used the utility function of upper confidence bound (UCB) [27] to inform which process conditions to acquire in the next batch. The second part is the probabilistic constraint for the acquisition function. The local penalization method[28] was used to select the batch of new data acquisitions. The idea is to generate an envelope function that reduces the acquisition value around the location of sample *n* in the batch when evaluating sample *n*+1. See SI Section 2.2 for the experimental cost analysis to determine the optimal batch size (*i.e.*, number of samples in each batch). We used the package of Bayesian Optimization in Python called *Emukit* [29], and the Gaussian Process package called *GPy* [30].

To incorporate the film quality information and the dataset from the relevant previous experiments, we defined two probabilistic constraints for the acquisition. By this method, we will sample fewer in the region where film quality is poor, or low PCE conditions in previous experiments. To be conservative, we further softened the probabilistic constraint (*i.e.*, Constraint Function 2) to a range of 0.8 – 1. The weighting factors were chosen to make the current data 5 times more important than the previous data. The knowledge from the previous data was transferred by multiplying the acquisition function with Constraint Function 2. Note that, because of the difference in scaling factor, Constraint Function 2 is much weaker than Constraint Function 1 (*i.e.*, 2.5 times difference). This difference reflects our perspective on how much the film quality and the previous data should influence the acquisition. The film quality should be much more relevant in this optimization than the previous device data.

## 3. Results and Discussion:

### 3.1 Sequential-Learning Process Optimization with Improved Efficiency

Fig. 2a plots the process optimization guided by the sequential learning framework and probabilistic constraints of visual inspection and previous experimental data. PCEs were measured in batches consisting of 20 conditions at a time to enable iteration and model feedback to suggest the subsequent round of process parameters. The highest PCE device of each process condition (the dark-green dots in Fig. 2a) was used in the optimization algorithm.

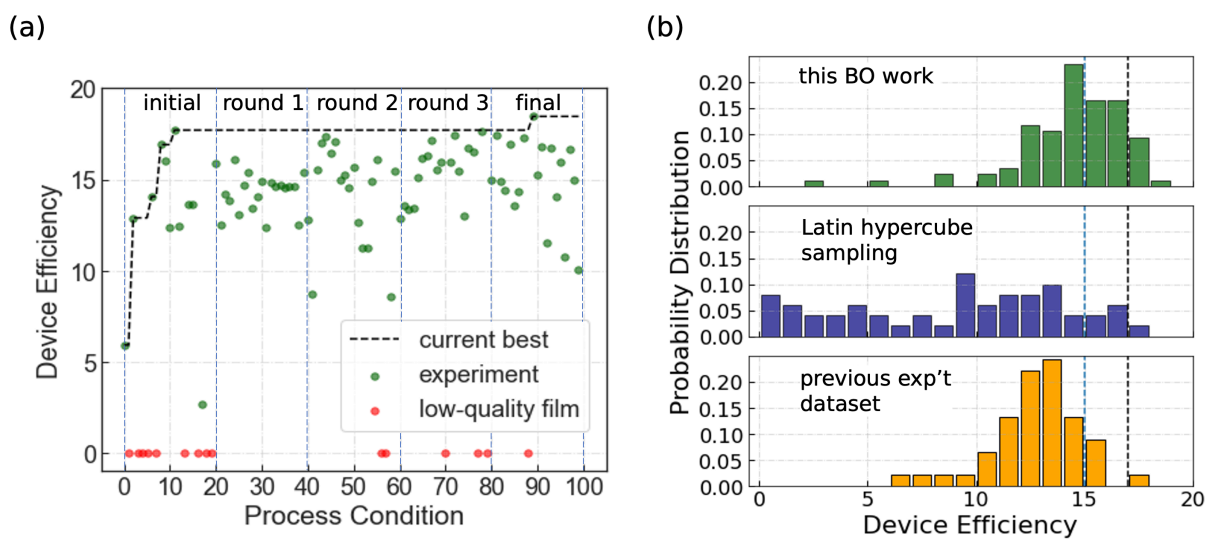

**Figure 2: Visualization of the experimental PCE data of solar cells with BO and other methods.** (a) The PCEs of the solar cell devices versus experimental process conditions. The solid dark-green dots are the experimental measurements of the highest efficiency among multiple cells at a given condition, and the solid red dots are low-quality films that did not pass the visual inspection. The black line is cumulatively the highest efficiency of all the devices until that specific process condition. (b) The distribution of power conversion efficiency for three different optimization workflows: BO-guided optimization in this work (85 conditions), Latin hypercube sampling (50 conditions), and a previous experimental dataset by researcher-led optimization (45 conditions). The dashed lines are marked for good performers at 15% PCE (blue) and top performers at 17% PCE (black).

In addition, based on the previous dataset of 45 process conditions (Fig. 2c), we defined a top performer with PCE to be a device with PCE above 17% and a good performer to be a device with PCE exceeding 15%. Thus, we had 1 top performer and 6 good performers. Among the 85 out of 100 conditions in BO-guided experiments in this work (excluding the low-quality films that did not pass the visual inspection), 45 process conditions achieved >15% PCE (good performers), and 10 process conditions achieved >17% PCE (top performers). The success rate was therefore 47% for the good performers and 12% for the top performers. In contrast, among the LHS-guided experiments of 50 process conditions (Fig. 2b), we found only 6 good performers (12% success rate), and 1 top performer (2% success rate).

Furthermore, the champion process condition produced a best-in-our-lab RSPP device efficiency of 18.5% in fewer than 100 conditions. The champion device in both LHS and previous experiments had never reached 18% PCE. According to a recent review paper on spray-deposited perovskite solar cells [31], this PCE is comparable to the highest-efficiency devices fabricated by spray deposition in the open air (18.5%) [32] and in the $N_2$ glovebox (19%) [33].

Visual inspection of the perovskite films was done after depositing the perovskite layer with RSPP. The film quality was rated by evaluating color, uniformity, and pinholes. The ratings were intended to be very conservative so that only the lowest-quality films were "tossed". This evaluation step is typical when optimizing with conventional researcher-driven device processing. Example low-quality and high-quality films are shown in Fig. S1.

To confirm the validity of our visual criteria, we fabricated additional devices from those low-quality films. We confirmed low PCE values for these conditions (*i.e.*, all below 13.5% with an average PCE of 7.8%), and the corresponding PCE distribution is shown in Fig. S2. This validation step confirms that the device fabrication for low-quality films can be skipped. When training our regression model with sequential learning, the film quality ratings were incorporated as a constraint function. Note that only 6 conditions after the first batch produced low-quality films, which further demonstrates that the model sampled the parameter space more effectively over time.

### 3.2 Incorporating Knowledge Constraints into Bayesian Optimization

Fig. 3 shows the machine learning method used in this study. Because a six-dimensional parameter space is difficult to visualize, two parameters of processing speed and substrate temperature were chosen as illustrations. A similar plot was obtained for each pair of parameters used in the regression analysis. A regression model was generated for the objective function and two constraint functions. The objective function contains the primary information measured from the experiments by plotting the process-efficiency relationship (Fig. 3a). The combination of probabilistic constraints are additional layers included based on visual inspection of film quality (Constraint Function 1, Fig. 3b) and previous experimental results (Constraint Function 2, Fig. 3c).

The resulting acquisition value was determined based on the selection criteria used (*i.e.*, PCE or film quality). These are plotted in Fig. 3d-3f and provide the framework for the sequential learning approach, namely how the model learned from each batch to suggest new process conditions with the goal of higher performance. The raw acquisition function in Fig. 3d was the upper confidence bound of the Objective Function. Probabilistic Constraint 1 is calculated based on the probability of a good-quality film, and Probabilistic Constraint 2 is calculated based on the probability of a device achieving the above-average PCE in a previous experimental dataset. Both probabilistic constraint functions were scaled to reduce the impact of these constraints and prevent the modification of the acquisition function from being too harsh.

More specifically, for Constraint Function 1, the film quality was rated in a binary outcome of either 0 (fail) or 1 (pass). The regression model of film quality was built to interpolate the unsampled regions (see Fig. 3e) and sequentially converted to the probability function of passing the film quality assessment. The probabilistic constraint function (*i.e.*, Constraint Function 1) was then softened to a range of 0.5 – 1 and multiplied by the raw acquisition function (Fig. 3b). Seentary Information (SI) Section 1.2 for the mathematical definition of probabilistic constraint function. The scaling means that we weigh the device data 2 times more important than the film quality data, to avoid any potential bias from our qualitative visual inspection of film quality.

The previous data (that can be found in Fig. 1b) for Constraint Function 2 was acquired during a preliminary optimization of the RSPP setup. We also use these data as a probabilistic constraint instead of a model prior because other variables beyond the six considered herein were slightly modified. For example, the spray and plasma nozzles used in previous experiments were different than in this work. Although we believe there is essential knowledge to be transferred (that is not affected too strongly by these modifications), the previous data are not equivalent to the current dataset. This required a more subtle way of representing the information as a probabilistic constraint to the BO framework, instead of directly adding the previous data in the model training.

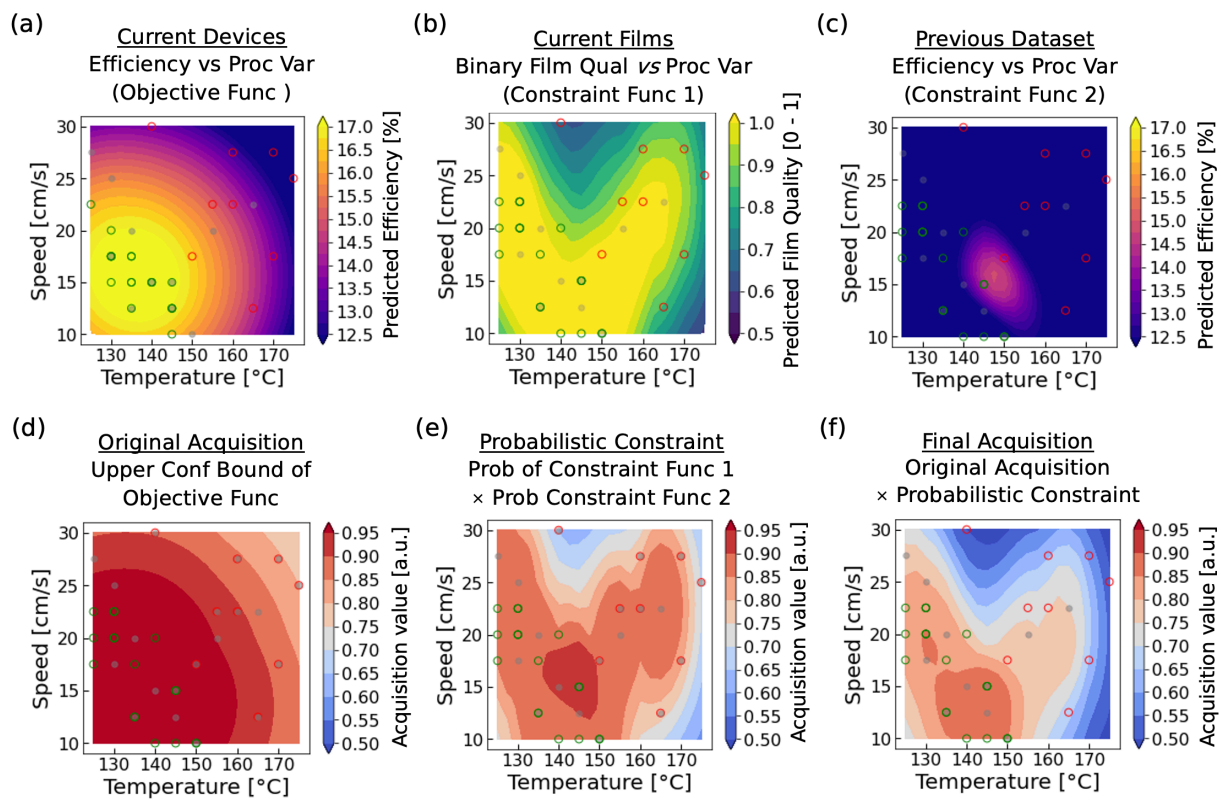

**Figure 3: A schematic illustration of probabilistic constraints for the acquisition function in the Bayesian optimization framework.** The proposed BO framework comprises the process-efficiency relation obtained from the current experimental data (Objective Function), the binary film quality evaluation (Constraint Function 1), and the process-efficiency relation from a previous study (Constraint Function 2). (a)-(c) are the outputs of the three regression models fitted to the respective data. (d)-(f) are the converted acquisition functions based on the trained regression models of (a)-(c).

### 3.3 Iterative Visualization of the Sample Distributions and Parameter Correlations

Fig. 4 visualizes the model learning through iteration as the acquisition function evolved and the process conditions began to converge round by round. Including the initial sampling by LHS, five experimental rounds were conducted for the device optimization. The experiments in Rounds 1-3 followed the suggestions from the BO acquisition function. Fig. 4 also shows that the probabilistic constraint began to affect the acquisition from Round 1. For example, the higher-temperature and higher-speed regions were sampled less after the initial sampling round because of the probabilistic constraint. This observation is consistent with the probabilistic constraint and acquisition function shown in Figs. 3e and 3f.

Due to the limited budget of 100 experimental conditions (or a total of 5 experimental rounds), we opted for a different acquisition method in the final round. Aiming at a further improvement of PCE, we conducted a local optimization in a smaller window of process conditions around the best condition predicted by the regression model. The best condition was found by the particle swarm optimization method [17,34], which is a common method for the global optimum for any given regression model. The parameter space of process conditions for the final round is shown in Table S2 in the SI.

The reason for this change is that our GP model has the tendency to "smooth out" features in the response surface as shown in Fig. 5. A similar finding was found in a previous study in materials science [10], where the GP fit was also very smooth to avoid potential overfitting to a small number of data points. To increase the probability of finding the optimum value, we use GP as a ringfencing technique to identify a region of the highest probability (*i.e.*, a window of process conditions), rather than a specific point or condition of highest probability. Within this region of highest probability, we combined a few techniques to achieve a balance of exploitation and exploration. Therefore, the final 20 conditions consisted of a best model-predicted condition, nearest neighbors of the best condition for exploitation, and LHS conditions in this region for more exploration.

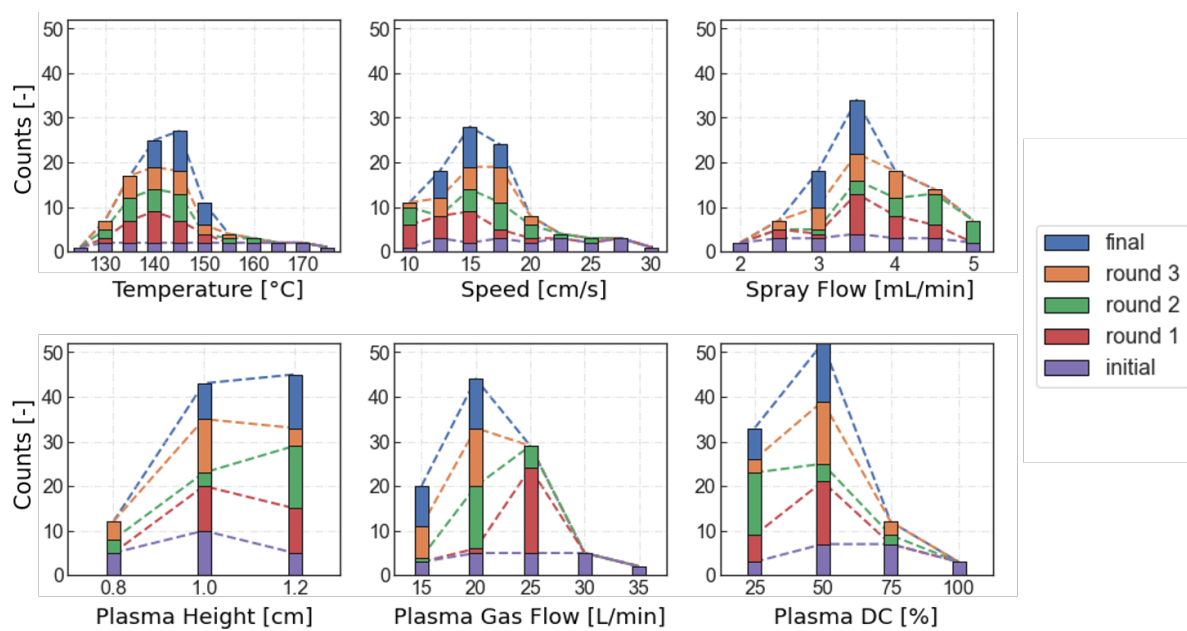

**Figure 4: The histogram distribution of process conditions used in each round of the optimization.** The initial samples used the LHS method. Rounds 1-3 were the process conditions suggested by Bayesian optimization and subsequently evaluated by experiments. The process converged to a small window after Round 3, so a sub-space with reduced parameter intervals was defined in the final-round optimization.

We visualized the trained regression model of process-efficiency relations by projecting the six-dimensional parameter space into 2D pair-wise contour plots. For each contour plot, we sampled two process variables of interest in a 20 x 20 grid. For every point in the contour plot (*i.e.*, a combination of the two variables), we sampled the remaining four variables 200 times, and we predicted the PCEs using the regression model for the 200 process conditions. Because we were interested in maximizing efficiency, we took the maximum PCE of the 200 process conditions. Therefore, these contour plots are the manifolds of the maximum PCEs in a 2D reduced space. For six process variables, there are a total number of 15 possible contour plots. Fig. 5 shows some examples of the dimension-reduced manifolds. For example, a data point in the contour plot indicates the combination of temperature = 140°C and plasma duty cycle = 20% can achieve a PCE of >16.5% when the other four process variables are fully optimized. Note that the predicted best efficiency by the trained regressor shown in the contour plots is lower than the experimentally measured PCE (*i.e.*, the ground truth) due to the training error of the regression model. This difference is common in Bayesian optimization, but it is still effective because an acquisition function balancing exploitation and exploration compensates for the model error. The training errors after each experimental iteration are found in Fig. S4 in the SI.

The manifold maps also inform correlations between different process variables and their impact on device efficiency. We also projected the suggested experimental conditions onto the contour plots, which helps interpret the decision-making of the algorithm and prevent mistakes at every round. Some correlations in Fig. 5 led to new insights of the RSPP method, including a negative correlation between both plasma gas flow/duty cycle and temperature (Fig. 5a and 5b). Since a certain curing dose is required to convert the precursor solution to a crystalline perovskite, this result indicated that the higher curing properties of increased plasma gas and duty cycle offset the lower temperature to balance the energy delivered to the perovskite during processing. Additionally, some correlative trends (consistent with previous experimentation) were also observed. For example, higher spray flow correlates with faster speed (see Fig. 5c), since a constant precursor dose should be delivered to the substrate to form the desired film thickness.

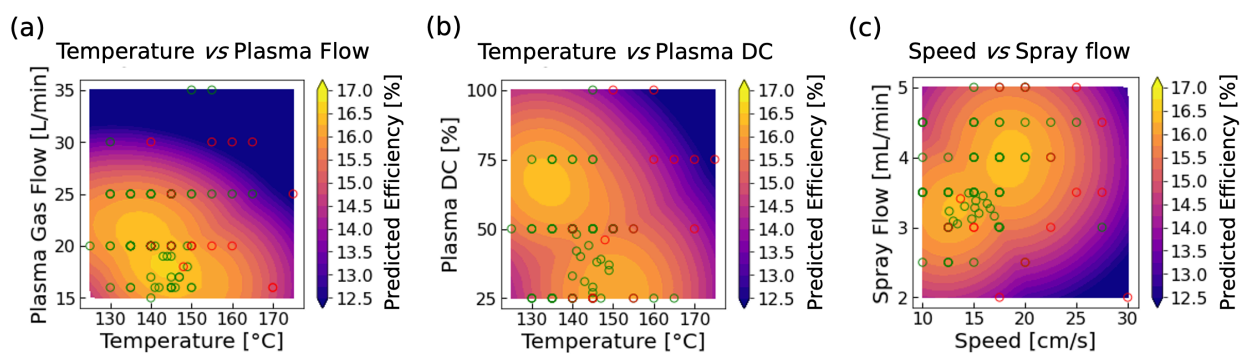

**Figure 5: Visualization of the process-efficiency relation based on the trained regression models.** The six-dimensional parameter space reduces into two-dimensional contour plots, and the highest efficiency for each condition pair is used for the contour plots. The green open circles are the process conditions made into devices. The red open circles are the process conditions made only into films that failed visual inspection. The color bars indicate the predicted mean efficiency from the trained regression model.

### 3.4 Benchmarking with Design-of-Experiment Methods using Virtual Optimization

We conducted a simulated "virtual" optimization with the fully trained "teacher" regression model for a benchmarking comparison between the sequential learning and conventional DoE methods. The "teacher" model is a Gradient Boosting regression model with decision trees, and it was trained with all the experimental data acquired for this work (*i.e.*, the data from the BO framework and the LHS shown in Fig. 2b). The training outcome is shown in Fig. 6a. For the "virtual" optimization, a good "teacher" model only needs to ensure the predictions follow a monotonic correlation with the ground truth (*i.e.*, the actual efficiency measured from experimentation). Therefore, following the study by Häse *et al.* [35], the spearman coefficient was used as an evaluation metric of the model training to assess the strength of the monotonic correlation between predicted and ground-truth efficiencies. This "teacher" regression model approximates the ground-truth manifold with a spearman coefficient of 0.93, and this model was used as a "teacher" model to run simulated optimization. The use of a simulation tool in benchmarking analysis avoids the time-consuming experimental work and has the advantage of generating better statistics for random or pseudo-random processes. However, because of the discrepancy of the absolute values between predictions and ground truths, we normalized the PCE values in the simulated optimization.

As mentioned previously, the full factorial sampling of all the process conditions in Table I is more than 40k samples, which is not feasible to be explored experimentally. The use of partial factorial Design of Experiment (DoE) methods (*e.g.*, generalized subset design [36], or D-optimal design [37]) could reduce the sample size to 1/2 or 1/3, but the number of samples is still too large (*i.e.*, greater than 10k). These methods are not fair comparisons with the BO framework. Alternatively, we used three model-free sampling methods: Latin hypercube sampling (LHS), one-variable-at-a-time sampling (OVATS) and factorial sampling with progressive grid subdivision (FS-PGS). The LHS, OVATS and FS-PGS methods are illustrated in detail in the Sections 3.2 –3.4 of SI. The results of different sampling methods are compared in Figs. 6b and 6c, together with the BO framework. Fig. 6b shows the PCE distributions for a limited experimental budget of 100 process conditions, whereas Fig. 6c shows the evaluation of the best PCE at the given number of process conditions.

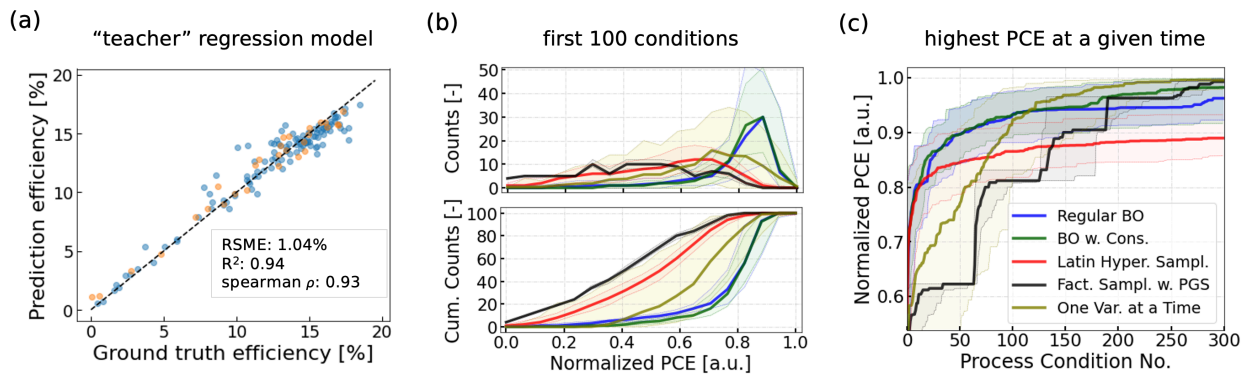

**Figure 6: Benchmarking of the newly developed sequential learning framework with the other design of experiment methods in the simulated process optimization.** (a) Prediction versus ground truth of the trained regression model of Gradient Boosting with Decision Trees. (b) The histogram distribution and the cumulative distribution of simulated PCEs of the first 100 process conditions plotted for different sampling methods. (c) The best PCEs at the given number of process conditions for different sampling methods, in the simulated optimizations until the 300th process condition. In (b) and (c): the sampling methods in the comparisons are regular BO (blue), BO with probabilistic constraints (dark green), Latin hypercube sampling (red), factorial sampling with progressive grid subdivision (black), and one-variable-at-at-time sampling (olive); the shaded red area represents the 5% − 95% confidence interval of an ensemble of 100 runs with random seeding; all the simulated PCEs are normalized by the highest simulated PCE.

First, Latin hypercube sampling is a popular model-free sampling method and produces a low-discrepancy pseudo-random sampling sequence (in contrast to the random sampling method). In this case, the outcome of the experiments does not inform the choice of the next round of samples. We ran 100 times and obtained the 5% – 95% confidence interval of the statistical distributions. LHS is one of the most commonly-used baseline comparison in BO studies [35], but it is not the best experimental practice in a research lab. Instead, we also implemented more realistic methods used in an experimental optimization, such as, OVATS and FS-PGS.

Second, one-variable-at-a-time sampling is one of the most popular methods for experimental planning by academic researchers [38]. Therefore, we believe that OVATS is suitable to benchmark ML methods against status quo. In this simulation, we started with one variable to screen and the other variables being constant, and we then fixed the optimal condition of the first variable and screened the second variable orthogonally. We kept iteratively screening the variables until the device efficiency did not further improve. When the efficiency stopped improving, we subdivided the grid and continued to screen more conditions per variables. The batch size varies with the number of levels for each variable. We randomized the choice of initial variable and the sampling sequence, and we ran the methods for 100 times to obtain the 5% – 95% confidence interval.

Third, factorial sampling with progressive grid subdivision (FS-PGS) is another method of a sequential design of experiment using factorial sampling, which is similar to the sampling concept proposed in Refs. [39] and [40]. Two-level grids of process variables were divided and sampled sequentially at every round in FS-PGS, and the parameter space was refined based on the process condition of the highest PCE found in the prior samples. Through this method, the process conditions were exploited locally around the highest efficiency with only a small fraction of the full factorial combinations. The batch size is 64 samples for the initial round (*i.e.*, $2^6$) and 63 samples subsequently (*i.e.*, $2^6$-1). We randomized the sample sequence for 100 times in each batch and produced the 5%-95% confidence interval.

To understand the manifold of the parameter space, we ran the full grid sampling methods and 100k random sampling with the "teacher" model. The PCE distributions are shown in Fig. S9 in SI. We found we were looking at a "needle-in-a-haystack" problem, where the top 5 percentile mark is at 0.8 in the normalized PCE, the top 1 percentile mark is at 0.85 in the normalized PCE, and the top 0.1 percentile mark is at 0.9 in the normalized PCE (as shown in Fig. S9c). We will use these marks to evaluate the performance of different experimental planning methods.

Fig. 6b compares the PCE distributions for different sampling methods given a limited experimental budget of 100 process conditions. We observe that both the regular BO and the BO with knowledge constraints achieved more high-efficiency devices in the early runs, and both were able to find 50 conditions at the top 1 percentile mark (*i.e.*, 0.8 in the normalized PCE). OVATS found 19 conditions exceeding 0.8 in the normalized PCE. LHS and FS-PGS has less than 5 conditions exceeding 0.8 in the normalized PCE. Therefore, in the task of finding the conditions of the top-one-percentile PCE, the BO methods have much higher probability of success, with an enhancement factor of 2.6x to OVATS, and more than 10x to LHS and FS-PGS.

The highest PCE versus the number of conditions are then compared in Fig. 6c. We see that, within the first 100 conditions, the BO methods performed significantly better in improving the PCE than OVATS, LHS, and FS-PGS. Because a few conditions were screened out by film quality as a probabilistic constraint at the first 20 conditions, BO with knowledge constraints was found to be slightly advantageous over the regular BO (and pure LHS).

In the comparison with OVATS and FS-PGS, we can see LHS is an efficient method for sampling, exceeding 0.8 in the normalized PCE (while OVATS and FS-PGS are below 0.7). However, the slope of improvement of LHS method flattened after the first 20 conditions. The BO methods reached 0.9 in the normalized PCE after 50 conditions, while LHS reached 0.85, and OVATS and FS-PGS were still below 0.8.

These observations in Fig. 6b and 6c indicate that the BO methods can help us to get closed to optimum very quickly during process optimization, because it exhibits an impressive acceleration at the small numbers of process conditions explored. However, the acceleration factor begins to decline at higher process conditions, indicating a decreasing rate of improvement of discovery of the highest PCE conditions. For example, at the 0.8 mark in the normalized PCE, the acceleration factor of the BO methods with LHS initial sampling is more than 5.5x over the OVATS method (*i.e.*, 55/10); at the 0.9 mark in the normalized PCE, the acceleration factor of the BO methods. is approximately 1.9x over the OVATS method (i.e., 95/50).

We also simulated the scenarios beyond the 100 conditions for a deeper analysis of different sampling methods. We observed the following points in Fig. 6c: (1) After 120 conditions, OVATS started overtaking the BO methods. This echoes the necessity of switching to a local search or an exploitative strategy in a later stage of the optimization. (2) the BO with knowledge constraint method started performing better than the regular BO method after the 160 conditions. (3) the FS-PGS method caught up to the BO methods after the 190 conditions. (4) OVATS and FS-PGS were able to get very closed the optimal PCE toward the end of the 300 conditions, while the BO with knowledge constraint method exceeds 0.98 in the normalized PCE, and the regular BO method exceeds 0.96 in the normalized PCE. (5) the OVATS has a significantly larger spread in the 5% – 95% confidence interval than the other methods, especially the lower bound of 5 percentile is much lower than the median for the first 100 conditions.

The decreasing rate of improvement in BO suggests that the future improvement of sequential learning may involve adaptive acquisition strategies, which switch between different acquisition methods based on what the algorithm deems to be advantageous. This adaptive acquisition may be implemented through combined acquisition functions or switching to another sampling method. The finding is consistent with our experimental optimization in this work, where the acquisition strategy was switched the final round to increase the chance of the PCE improvement. The first attempt towards this has been implemented in the optimization study of Ag nanoparticle fabrication [10], where a neural network (NN) model was run in parallel with the GP, to improve regression resolution near the region of maximum interest. In that study, however, the NN was not used to inform the acquisition function, as would be the case here.

### 4. Conclusions:

In this work, a new sequential learning framework—Bayesian optimization with knowledge constraints—was used for the process optimization of perovskite solar cells, which intelligently incorporated the previous data from preliminary optimization experiments and researchers' visual evaluation of perovskite film quality. The workflow of the BO framework with knowledge constraints effectively mimics the "conventional" iterative optimization method and allows for the flexible incorporation of multiple information sources during process optimization. Compared to regular BO, the concept of knowledge constraints shows two key advantages: (1) it provides a flexible way to incorporate the knowledge learned in the past (or qualitative inputs from an experienced researcher) and (2) it reduces the extra experimental efforts of converting the film into devices when the film quality is visibly poor.

With the experimental planning guided by the Bayesian optimization framework with knowledge constraints, we achieved an 18.5%-efficient solar cell in open-air after optimizing six process variables for perovskite deposition, conducting 5 experimental rounds and screening 100 process conditions. This is the highest PCE achieved to date with the RSPP method and done with only a very small fraction of the more than 40k possible combinations in a grid search (also known as full factorial sampling). In addition to the best-in-our-lab PCE values, we also discovered 10 process conditions that produced ≥17% PCE devices with BO, significantly outperforming the pseudo-random sampling method of LHS in a parallel run where only 1 out of 50 process conditions was above 17% PCE. This indicates a >5x enhancement in finding the process conditions producing the top-performing solar cells.

To benchmark the performance of the BO frameworks (*i.e.*, both regular BO and BO with knowledge constraints), we trained a teacher regression model to produce the "ground truth" (i.e., simulated experiment results), and we ran virtual optimizations with three common sampling methods for experimental planning, namely Latin hypercube sampling, one-variable-at-a-time sampling method, and factorial sampling with progressive grid subdivision. We also found that both BO methods significantly overperformed the OVATS method within the first 100 process conditions, but the OVATS method overtook the BO methods afterward. This shows the true advantage of the BO framework is getting close to the optimum quickly within a small budget of experiments, and further development of adaptive acquisition strategies (*e.g.,* time-evolving acquisition function) might be needed.

In summary, this framework of BO with knowledge constraints could be widely applied in process optimization and materials screening for perovskite PV devices, regardless of deposition methods (*e.g.*, spin coating, blade coating, slot-die coating, or thermal evaporation). Moreover, the knowledge constraint is also flexible to incorporate quantitative film quality measurements, such as, light-beam induced current mapping, photoluminescence imaging and quasi-fermi level splitting mapping [41]. This will be particularly helpful to tackle the challenge of perovskite PV scale-up by incorporating the learning from lab-scale devices into the optimization of large-area modules.

**Contributions:**

Z.L., N.R., R.H.D. and T.B. contributed to the project ideation, the initial planning, the data analysis, and the paper revision. Z.L. led the development and execution of the sequential learning framework with probabilistic constraints, with Z.R.'s contributions. N.R. led the experimental fabrication and device characterization, with the contributions from A.C.F and T.C. Z.L and N.R. co-led the project coordination, and the manuscript writing (with the inputs from all the co-authors). R.H.D and T.B. provided guidance during the project execution and the manuscript writing.

**Conflict of Interest Statement:**

Some of the authors (Z.L. and T.B.) hold equity in a start-up company, Xinterra, focused on commercializing machine learning technologies for accelerated materials development. To ensure equal access for all, materials from this study (*e.g.*, code, datasets) are open sourced.

**Data and code availability**

The codes and the datasets used for conducting Bayesian optimization with probabilistic constraints are available in GitHub repository: https://github.com/PV-Lab/SL-PerovskiteOpt.

**Acknowledgments:**

We acknowledge the discussion on machine learning methods with Dr. Felipe Oviedo, who was a former member of MIT PVLab and is now at Microsoft AI for Good. The authors thank Shaffiq Jaffar from TotalEnergies SE for his feedback for this research work, and the general discussions about machine learning in energy-materials research. Z.L. and T.B. thank the feedback from Dr. Shijing Sun, Dr. Armi Tiihonen and other members from the MIT PVLab. Z.L. thanks the *Emukit* development team for fruitful discussions on the GitHub forum.

This material is based upon the work supported by the U.S. Department of Energy's Office of Energy Efficiency and Renewable Energy (EERE) under the Solar Energy Technology Office (SETO) Award Number DE-0009366. N.R. acknowledges partial support from the U.S. Department of Energy's Award Number DE-EE0008559. Z.L. acknowledges partial support from a TotalEnergyies SEFellowship from the MIT Energy Initiative. T.C. acknowledges the support from the Graduate Research Fellowship Program (Award No. DGE-656518) from the U.S. National Science Foundation. Z.R. acknowledges the support from Singapore-MIT Alliance for Research and Technology (SMART), particularly its Independent Research Group of "Low Energy Electronic Systems". SMART is established by the MIT in partnership with the National Research Foundation (NRF) of Singapore.

**References:**

1. Li, Z., Klein, T.R., Kim, D.H., Yang, M., Berry, J.J., Van Hest, M.F.A.M., and Zhu, K. (2018). Scalable fabrication of perovskite solar cells. Nat. Rev. Mater. *3*, 1–20.

2. Almora, O., Baran, D., Bazan, G.C., Berger, C., Cabrera, C.I., Catchpole, K.R., Erten-Ela, S., Guo, F., Hauch, J., Ho-Baillie, A.W.Y., et al. (2020). Device Performance of Emerging Photovoltaic Materials (Version 1). Adv. Energy Mater. *11*, 2002774.

3. Yoo, J.J., Seo, G., Chua, M.R., Park, T.G., Lu, Y., Rotermund, F., Kim, Y.-K., Moon, C.S., Jeon, N.J., Correa-Baena, J.-P., et al. (2021). Efficient perovskite solar cells via improved carrier management. Nature *590*, 587–593.

4. Rolston, N., Scheideler, W.J., Flick, A.C., Chen, J.P., Elmaraghi, H., Sleugh, A., Zhao, O., Woodhouse, M., and Dauskardt, R.H. (2020). Rapid Open-Air Fabrication of Perovskite Solar Modules. Joule *4*, 2675–2692.

5. Burger, B., Maffettone, P.M., Gusev, V. V, Aitchison, C.M., Bai, Y., Wang, X., Li, X., Alston, B.M., Li, B., Clowes, R., et al. (2020). A mobile robotic chemist. Nature *583*, 237–241.

6. Granda, J.M., Donina, L., Dragone, V., Long, D.L., and Cronin, L. (2018). Controlling an organic synthesis robot with machine learning to search for new reactivity. Nature *559*, 377–381.

7. Shields, B.J., Stevens, J., Li, J., Parasram, M., Damani, F., Alvarado, J.I.M., Janey, J.M., Adams, R.P., and Doyle, A.G. (2021). Bayesian reaction optimization as a tool for chemical synthesis. Nature *590*, 89–96.

8. Gongora, A.E., Xu, B., Perry, W., Okoye, C., Riley, P., Reyes, K.G., Morgan, E.F., and Brown, K.A. (2020). A Bayesian experimental autonomous researcher for mechanical design. Sci. Adv. *6*, eaaz1708.

9. MacLeod, B.P., Parlane, F.G.L., Morrissey, T.D., Häse, F., Roch, L.M., Dettelbach, K.E., Moreira, R., Yunker, L.P.E., Rooney, M.B., Deeth, J.R., et al. (2020). Self-driving laboratory for accelerated discovery of thin-film materials. Sci. Adv. *6*, eaaz8867.

10. Mekki-Berrada, F., Ren, Z., Huang, T., Wong, W.K., Zheng, F., Xie, J., Tian, I.P.S., Jayavelu, S., Mahfoud, Z., Bash, D., et al. (2021). Two-step machine learning enables optimized nanoparticle synthesis. npj Comput. Mater. *7*, 55.

11. Attia, P.M., Grover, A., Jin, N., Severson, K.A., Markov, T.M., Liao, Y.H., Chen, M.H., Cheong, B., Perkins, N., Yang, Z., et al. (2020). Closed-loop optimization of fast-charging protocols for batteries with machine learning. Nature *578*, 397–402.

12. Lookman, T., Balachandran, P. V, and Xue, D. (2019). Active learning in materials science with emphasis on adaptive sampling using uncertainties for targeted design. npj Comput. Mater. *5*, 21.

13. Nikolaev, P., Hooper, D., Webber, F., Rao, R., Decker, K., Krein, M., Poleski, J., Barto, R., and Maruyama, B. (2016). Autonomy in materials research: A case study in carbon nanotube growth. npj Comput. Mater. *2*, 1–6.

14. Balachandran, P. V, Kowalski, B., Sehirlioglu, A., and Lookman, T. (2018). Experimental search for high-temperature ferroelectric perovskites guided by two-step machine learning. Nat. Commun. *9*, 1–9.

15. Ling, J., Hutchinson, M., Antono, E., Paradiso, S., and Meredig, B. (2017). High-Dimensional Materials and Process Optimization using Data-driven Experimental Design with Well-Calibrated Uncertainty Estimates. Integr. Mater. Manuf. Innov. *6*, 207–217.

16. Rohr, B., Stein, H.S., Guevarra, D., Wang, Y., Haber, J.A., Aykol, M., Suram, S.K., and Gregoire, J.M. (2020). Benchmarking the acceleration of materials discovery by sequential learning. Chem. Sci. *11*, 2696–2706.

17. James V. Miranda, L. (2018). PySwarms: a research toolkit for Particle Swarm Optimization in Python. J. Open Source Softw. *3*, 433.

18. Wang, Z., Gehring, C., Kohli, P., and Jegelka, S. (2018). Batched large-scale Bayesian optimization in high-dimensional spaces. In Proceedings of the 21st International Conference on Artificial Intelligence and Statistics (AISTATS), pp. 745–754.

19. Harris, S.J., Harris, D.J., and Li, C. (2017). Failure statistics for commercial lithium ion batteries: A study of 24 pouch cells. J. Power Sources *342*, 589–597.

20. Ziatdinov, M., Ghosh, A., and Kalinin, S. V (2022). Physics makes the difference: Bayesian optimization and active learning via augmented Gaussian process. Mach. Learn. Sci. Technol. *Just Accep*.

21. Gelbart, M.A., Snoek, J., and Adams, R.P. Bayesian Optimization with Unknown Constraints.

22. Sun, S., Tiihonen, A., Oviedo, F., Liu, Z., Thapa, J., Zhao, Y., Hartono, N.T.P., Goyal, A., Heumueller, T., Batali, C., et al. (2021). A data fusion approach to optimize compositional stability of halide perovskites. Matter *4*, 1305–1322.

23. Hilt, F., Hovish, M.Q., Rolston, N., Brüning, K., Tassone, C.J., and Dauskardt, R.H. (2018). Rapid route to efficient, scalable, and robust perovskite photovoltaics in air. Energy Environ. Sci. *11*, 2102–2113.

24. Scheideler, W.J., Rolston, N., Zhao, O., Zhang, J., and Dauskardt, R.H. (2019). Rapid Aqueous Spray Fabrication of Robust NiO$_x$: A Simple and Scalable Platform for Efficient Perovskite Solar Cells. Adv. Energy Mater. *9*, 1803600.

25. Lim, Y.-F., Ng, C.K., Vaitesswar, U.S., and Hippalgaonkar, K. (2021). Extrapolative Bayesian Optimization with Gaussian Process and Neural Network Ensemble Surrogate Models. Adv. Intell. Syst. *3*, 2100101.

26. Liang, Q., Gongora, A.E., Ren, Z., Tiihonen, A., Liu, Z., Sun, S., Deneault, J.R., Bash, D., Mekki-Berrada, F., Khan, S.A., et al. (2021). Benchmarking the performance of Bayesian optimization across multiple experimental materials science domains. npj Comput. Mater. 2021 71 *7*, 118.

27. Niranjan Srinivas, Andreas Krause, Sham M. Kakade, and Matthias W. Seeger (2012). Information-Theoretic Regret Bounds for Gaussian Process Optimization in the Bandit Setting. IEEE Trans. Inf. Theory *58*, 3250–3265.

28. González, J., Dai, Z., Hennig, P., and Lawrence, N. (2016). Batch Bayesian Optimization via Local Penalization. In Proceedings of Machine Learning Research, pp. 648–657.

29. Paleyes, A., Pullin, M., Mahsereci, M., Lawrence, N., and González, J. (2019). Emulation of physical processes with Emukit. In NeurIPS Workshop on Machine Learning and the Physical Sciences, p. Online Available: https://ml4physicalsciences.gith.

30. GPy: A Gaussian process framework in python (2012).

31. Bishop, J.E., Smith, J.A., and Lidzey, D.G. (2020). Development of Spray-Coated Perovskite Solar Cells. ACS Appl. Mater. Interfaces *12*, 48237–48245.

32. Su, J., Cai, H., Yang, J., Ye, X., Han, R., Ni, J., Li, J., and Zhang, J. (2020). Perovskite Ink with an Ultrawide Processing Window for Efficient and Scalable Perovskite Solar Cells in Ambient Air. ACS Appl. Mater. Interfaces *12*, 3531–3538.

33. Ding, J., Han, Q., Ge, Q.Q., Xue, D.J., Ma, J.Y., Zhao, B.Y., Chen, Y.X., Liu, J., Mitzi, D.B., and Hu, J.S. (2019). Fully Air-Bladed High-Efficiency Perovskite Photovoltaics. Joule *3*, 402–416.

34. Kennedy, J., and Eberhart, R. (1995). Particle swarm optimization. Proc. Int. Conf. Neural Networks *4*, 1942–1948.

35. Häse, F., Aldeghi, M., Hickman, R.J., Roch, L.M., Christensen, M., Liles, E., Hein, J.E., and Aspuru-Guzik, A. (2021). Olympus: a benchmarking framework for noisy optimization and experiment planning. Mach. Learn. Sci. Technol. *2*, 035021.

36. Surowiec, I., Vikström, L., Hector, G., Johansson, E., Vikström, C., and Trygg, J. (2017). Generalized Subset Designs in Analytical Chemistry. Anal. Chem. *89*, 6491–6497.

37. NIST/SEMATECH e-Handbook of Statistical Methods 5.5.2.1. D-Optimal designs https://www.itl.nist.gov/div898/handbook/pri/section5/pri521.htm.

38. Cao, B., Adutwum, L.A., Oliynyk, A.O., Luber, E.J., Olsen, B.C., Mar, A., and Buriak, J.M. (2018). How to optimize materials and devices via design of experiments and machine learning: Demonstration using organic photovoltaics. ACS Nano *12*, 7434–7444.

39. Kurchin, R., Romano, G., and Buonassisi, T. (2019). Bayesim: A tool for adaptive grid model fitting with Bayesian inference. Comput. Phys. Commun. *239*, 161–165.

40. Eriksson, D., Ai, U., Pearce, M., Gardner, J.R., Turner, R., and Poloczek, M. (2019). Scalable Global Optimization via Local Bayesian Optimization. In Proceedings of the 33rd International Conference on Neural Information Processing Systems (NIPS), pp. 5496–5507.

41. Stolterfoht, M., Wolff, C.M., Márquez, J.A., Zhang, S., Hages, C.J., Rothhardt, D., Albrecht, S., Burn, P.L., Meredith, P., Unold, T., et al. (2018). Visualization and suppression of interfacial recombination for high-efficiency large-area pin perovskite solar cells. Nat. Energy 2018 310 *3*, 847–854.

# Machine Learning with Knowledge Constraints for Process Optimization of Open-Air Perovskite Solar Cell Manufacturing

Zhe Liu[1,*,§], Nicholas Rolston[2,*], Austin C. Flick[2], Thomas W. Colburn[2], Zekun Ren[3],

Reinhold H. Dauskardt[2,†], Tonio Buonassisi[1,†]

[1]Massachusetts Institute of Technology, Cambridge, MA, United States
[2]Stanford University, Stanford, CA, United States
[3]Singapore-MIT Alliance for Research and Technology, Singapore

*These authors contributed equally as co-first authors.
†Email correspondence: buonassi@mit.edu (T.B.) & rhd@stanford.edu (R.H.D)
§This author is now at Northwestern Polytechnical University (NPU), Xi'an, Shaanxi, P.R. China

# Supplementary Information

***Table of Contents***

# 1. Sequential Learning Framework to Guide Iterative Learning Cycles

The sequential learning contains five steps in each iterative cycle as discussed in the main text. In addition, some key features used in this work has been summarized as follows.

- Initial sampling strategy: For a general framework, we start with experimental planning of process conditions with a Latin-hypercube sampling (LHS) method for the initial round. LHS offers good coverage of the entire parameter space, with as few shots on goal as possible (because experiments are expensive), yielding high information at the start of the learning.
- Manufacturing & testing: Perovskite solar cells are fabricated by the RSPP method at Stanford, and PCE is measured with a solar simulator under standard testing conditions.
- Model training: With the experimental data of the process parameters and the device PCEs, we train the regression model to learn the process-efficiency relation. The surrogate model selected was Gaussian Process (GP) regression with anisotropic Matern 5/2 kernel.
- Predicting & planning: The prediction results were evaluated by an acquisition function together with the constraint information, and therefore a new round of experiments was planned using Bayesian optimization (BO). We explain some details of the constraint information by film quality and the acquisition function choice in this section.

*1.1 Perovskite Film Quality as Probabilistic Constraint*

In the proposed BO framework in this work, visual assessment has been used a quick method to create a probabilistic constraint in the acquisition decisions. Fig. S1 shows the examples for low-quality and good-quality films with visual inspection. The low-quality film has a pale gray color on the edges (which is an indication of poor crystallization kinetics and a porous film). This low-quality film is also not very uniform and has many large holes. In contrast, the good film has a uniform brown color across the substrate.

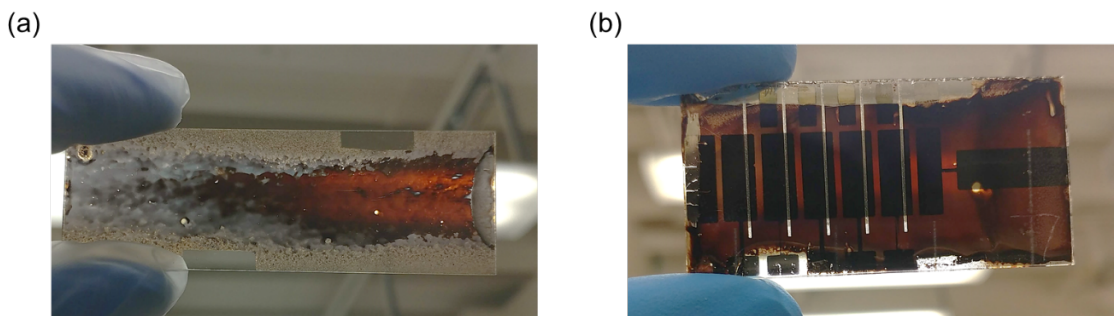

Figure S1: The pictures of (a) a poor-quality film that failed the visual inspection, and (b) a good-quality film that were fabricated into devices.

To validate the film quality assessment, we made the low-quality films into devices and measured the power conversion efficiencies. The histogram of these device efficiencies is shown in Fig. S2, and they are all in the low-efficiency regime. None is above 15% efficiency.

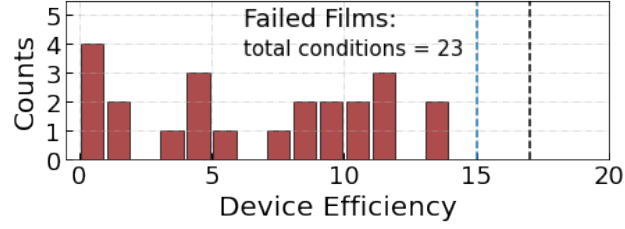

Figure S2: The efficiency distribution for those devices that were fabricated using low-quality films (as a confirmation of the visual inspection). The blue dashed line marks 15% efficiency, and the black dashed line marks 17% efficiency.

*1.2 Choice of Acquisition and Constraint Functions*

a. Upper Confidence Bound

For the acquisition function of the objective function (*i.e.*, process-efficiency relation), we used the commonly-used utility function named upper confidence bound (UCB) [1], which is also known as Negative Lower Confidence Bound (LCB) in the *Emukit* package and other literature. UCB achieves a better balance between exploration and exploitation in the acquisition process, in comparison to the other popular utility function named Expected Improvement [2]. The UCB function is defined as,

$$acq_{\text{UCB}} = y_{\text{pred}} + \beta \cdot \sigma_{\text{pred}}$$

where $\beta$ is the parameter that adjusts the relative weight to prediction uncertainty $\sigma_{\text{pred}}$ over the prediction mean $y_{\text{pred}}$. In the term of balancing exploration and exploitation, higher $\beta$ value leads to more exploration, while lower $\beta$ value leads to more exploitation. In this work, we set $\beta$ equals to 1, because we found $\beta$ = 1 achieved a good balance in the previous works [3].

b. Probability of Satisfying the Constraint Criteria

For converting the film quality model and the regression model of the previous device data, we calculate the probability of exceeding a threshold value $\delta$:

$$pr_{\text{cons}} = \text{CDF}\left(\frac{y_{\text{cons}} - \delta}{\sigma_{\text{cons}}}\right)$$

where CDF is the cumulative distribution function of normal distribution.

## 2. Apply the Sequential Learning to RSPP Experiments

*2.1 Five Cycles of Learning*

We have listed the different sampling methods in Table S1 for the different batches (*i.e.*, 5 experimental batches in total of 20 process conditions). The initial round and final round were guided by global LHS and local region search respectively (as explained in the main text). We recorded the relative humidity (RH) changes in the process chamber in the table to track a potential source for the variability, although we demonstrated the PCEs of RSPP devices were fairly independent of ambient RH in [4].

Table S1: Relative humidity in the RSSP deposition chamber for each batch of devices.

|  | **Initial** | **Round 1** | **Round 2** | **Round 3** | **Final** |
|---|---|---|---|---|---|
| Process Condition | 0 – 19 | 20 – 39 | 40 – 59 | 60 – 79 | 80 – 99 |
| Acquisition Method | LHS | BO with UCB | BO with UCB | BO with UCB | Local region search |
| Relative Humidity [%] | 43% | 27% | 25% | 50% | 35% |
| The ambient temperature is 23°C±2°C. | | | | | |

Table S2: The parameter space for the final round of the local region search.

| Process Variable | Total Range (Interval) | Process Variable | Total Range (Interval) |
|---|---|---|---|
| Temperature | 140 – 150 °C (1°C) | Plasma height | 1.0 – 1.2 cm (0.05 cm) |
| Speed | 12.5 – 17.5 cm/s (0.1 cm/s) | Plasma gas flowrate | 16 – 20 L/min (1 L/min) |
| Spray flowrate | 3 – 3.5 mL/s (0.01 mL/min) | Plasma duty cycle | 25 – 50 % (1%) |

*2.2 Choice of Batch Size in Each Cycle*

During sequential learning, the batch size in every iteration is an importance parameter to choose for Bayesian optimization. Although smaller batches can provide quicker feedback to revise the regression model, it could be much more costly (in term of time consumption for the experiments) since many processes are batch process. We estimated the time consumption for the device fabrication and listed them in Table S3. We only listed the key steps that consumes the most time; other steps, such as precursor preparation and data analysis, can be added as well.

Based on Table S3, we calculated the time needed for each batch and average time per substrate while varying the batch size. Fig. S3 gives a quantitative perspective of time consumption per batch and average time per substrate for different batch size. Especially observed in Fig. S3b, the average time per substrate is diminishing very quickly with increasing batch size. That is why the batch mode of Bayesian optimization (with local penalization algorithm) is more cost-effective than single-sample iteration when experiments are more time-consuming than computation. A batch size is chosen to achieve a good balance between low average time cost per substrate and the batch time to feedback the regression model. Because the most time-consuming step – thermal evaporation – can accommodate up to 20 substrates per batch, we used the batch size of 20 in this work. Note the sudden jump in both Fig. S3a and S3b at the batch size of 21.

Table S3: Estimated time consumption for some key steps in device fabrication.

| Steps | Setup Time | Processing Time |
| --- | --- | --- |
| Substrate Cleaning | 10 mins to fill cleaning agents | 50 mins for up to 20 substrates |
| RSPP Deposition | 1hr to reconfigure & test the tool | 10 – 15 mins per condition |
| Evaporation (batch-20) | 30 mins to pump the pressure down | 3 hrs. for up to 20 substrates |
| *I-V* characterization | 15 mins to warm up the lamp + circuit setup | 10 mins per substrate |

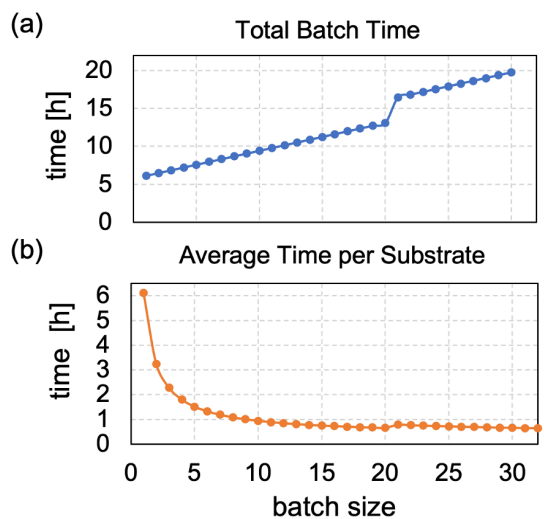

Figure S3: (a) The total time per batch for different batch size. (b) The average time per substrate for different batch size.

2.3 Evaluation of Surrogate Regression Model

After every experimental batch being made, the new data were added to the training dataset to revise the regression model. Fig. S4 shows the model prediction vs the ground truth (*i.e.*, the experimentally measured) at each iteration.

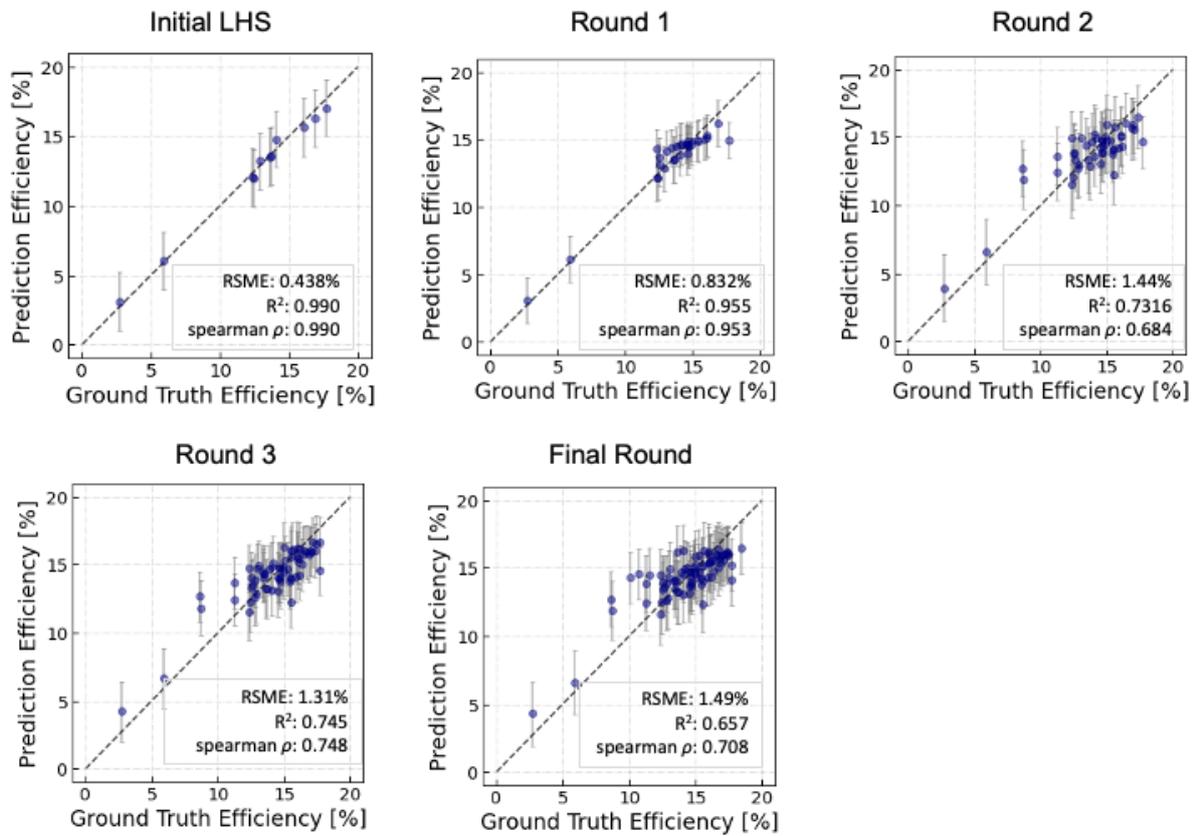

Figure S4: The predictions by surrogate model versus the ground truth of the measured PCEs after every experimental iteration. The error bars are the estimated standard deviation for the prediction uncertainty (*i.e.*, 68.2% confidence interval). The numerical metric for the model errors (i.e., root mean squared error RMSE, goodness of fit $R^2$, spearman coefficient) are shown in the inserted box.

2.4 Characterization Results of Perovskite PV Devices

a. All sampled experimental process conditions and solar cell device efficiencies

Table S4. All the sampled experimental process conditions and the fabricated device efficiency.

| ML Condition | Temperature [°C] | Speed [mm/s] | Spray Flow [uL/min] | Plamsa Height [cm] | Plasma Gas Flow [L/min] | Plasma DC [%] | Efficiency [%] | Film Success or not? |
|---|---|---|---|---|---|---|---|---|
| 0 | 155 | 200 | 5000 | 1.2 | 35 | 50 | 5.91 | Yes |
| 1 | 160 | 225 | 3000 | 0.8 | 30 | 75 | 0.06 | No |
| 2 | 135 | 200 | 2500 | 1 | 25 | 75 | 12.89 | Yes |
| 3 | 150 | 175 | 2000 | 1 | 20 | 100 | 4.77 | No |
| 4 | 170 | 275 | 4500 | 1.2 | 16 | 50 | 11.82 | No |
| 5 | 175 | 250 | 3500 | 1.2 | 25 | 75 | 9.06 | No |
| 6 | 140 | 150 | 4000 | 1 | 20 | 75 | 14.05 | Yes |
| 7 | 155 | 225 | 4000 | 0.8 | 30 | 25 | 9.88 | No |
| 8 | 130 | 175 | 3500 | 0.8 | 25 | 75 | 16.90 | Yes |
| 9 | 135 | 125 | 2500 | 1.2 | 20 | 25 | 16.05 | Yes |
| 10 | 130 | 250 | 4500 | 1.2 | 30 | 50 | 12.34 | Yes |
| 11 | 145 | 125 | 3500 | 1 | 25 | 50 | 17.70 | Yes |
| 12 | 145 | 150 | 4500 | 1 | 16 | 100 | 12.43 | Yes |
| 14 | 165 | 225 | 4000 | 1 | 25 | 25 | 13.66 | Yes |
| 15 | 125 | 275 | 3000 | 0.8 | 20 | 50 | 13.61 | Yes |
| 16 | 170 | 175 | 5000 | 1 | 16 | 75 | 13.40 | No |
| 17 | 150 | 100 | 2500 | 1 | 35 | 50 | 2.70 | Yes |
| 18 | 140 | 300 | 2000 | 1 | 30 | 50 | - | No |
| 19 | 165 | 125 | 3000 | 1 | 30 | 75 | 11.42 | No |
| 20 | 130 | 200 | 4500 | 1.2 | 25 | 50 | 15.87 | Yes |
| 21 | 135 | 125 | 3500 | 1 | 25 | 25 | 12.54 | Yes |
| 22 | 135 | 150 | 2500 | 1.2 | 20 | 25 | 14.16 | Yes |
| 23 | 135 | 150 | 3500 | 1.2 | 25 | 25 | 13.84 | Yes |
| 24 | 135 | 175 | 4000 | 1.2 | 25 | 25 | 16.12 | Yes |
| 25 | 135 | 175 | 4000 | 1.2 | 25 | 50 | 13.04 | Yes |
| 26 | 140 | 100 | 3500 | 1 | 25 | 50 | 14.72 | Yes |
| 27 | 140 | 100 | 3500 | 1.2 | 25 | 25 | 15.38 | Yes |
| 28 | 140 | 100 | 4000 | 1 | 25 | 50 | 13.41 | Yes |
| 29 | 140 | 125 | 2500 | 1 | 25 | 50 | 14.09 | Yes |
| 30 | 140 | 125 | 3500 | 1.2 | 25 | 25 | 14.89 | Yes |
| 31 | 140 | 150 | 3500 | 1 | 25 | 50 | 12.34 | Yes |
| 32 | 140 | 150 | 4000 | 1 | 25 | 50 | 14.85 | Yes |
| 33 | 145 | 100 | 3500 | 1.2 | 25 | 50 | 14.61 | Yes |
| 34 | 145 | 100 | 4500 | 1 | 25 | 50 | 14.67 | Yes |
| 35 | 145 | 125 | 3000 | 1 | 25 | 50 | 14.54 | Yes |
| 36 | 145 | 150 | 3500 | 1 | 25 | 50 | 14.61 | Yes |
| 37 | 145 | 150 | 4000 | 1 | 25 | 50 | 14.61 | Yes |
| 38 | 150 | 125 | 3500 | 1.2 | 25 | 50 | 12.54 | Yes |
| 39 | 150 | 150 | 4500 | 1.2 | 25 | 50 | 15.41 | Yes |
| 40 | 130 | 150 | 5000 | 1.2 | 25 | 25 | 12.82 | Yes |
| 41 | 130 | 225 | 4500 | 1.2 | 20 | 25 | 8.72 | Yes |
| 42 | 135 | 100 | 4500 | 1.2 | 20 | 25 | 15.50 | Yes |
| 43 | 135 | 150 | 3500 | 1.2 | 20 | 25 | 16.97 | Yes |
| 44 | 135 | 150 | 4000 | 0.8 | 20 | 50 | 17.37 | Yes |
| 45 | 135 | 175 | 3000 | 0.8 | 20 | 75 | 16.42 | Yes |
| 46 | 135 | 175 | 4000 | 1.2 | 20 | 50 | 17.04 | Yes |
| 47 | 140 | 100 | 3500 | 1 | 15 | 25 | 14.97 | Yes |
| 48 | 140 | 175 | 4500 | 1 | 20 | 25 | 15.22 | Yes |
| 49 | 140 | 175 | 5000 | 1.2 | 20 | 25 | 14.55 | Yes |
| 50 | 140 | 200 | 4000 | 0.8  | 20 | 75 | 15.65 | Yes |
| 51 | 140 | 200 | 5000 | 1.2  | 20 | 50 | 12.61 | Yes |
| 52 | 145 | 100 | 4500 | 1.2  | 20 | 25 | 11.23 | Yes |
| 53 | 145 | 100 | 4500 | 1.2  | 25 | 25 | 11.25 | Yes |
| 54 | 145 | 175 | 3500 | 1    | 20 | 50 | 14.93 | Yes |
| 55 | 145 | 175 | 4000 | 1.2  | 20 | 25 | 16.08 | Yes |
| 56 | 145 | 200 | 5000 | 1.2  | 25 | 25 | 11.18 | No  |
| 57 | 145 | 200 | 5000 | 1.2  | 25 | 25 | 10.68 | No  |
| 58 | 155 | 150 | 4500 | 1.2  | 25 | 25 | 8.58  | Yes |
| 59 | 160 | 150 | 4500 | 1.2  | 25 | 25 | 15.49 | Yes |
| 60 | 130 | 125 | 4000 | 1    | 20 | 25 | 12.84 | Yes |
| 61 | 130 | 175 | 3000 | 0.8  | 16 | 50 | 13.55 | Yes |
| 62 | 135 | 100 | 3500 | 1    | 16 | 50 | 13.38 | Yes |
| 63 | 135 | 125 | 2500 | 1    | 20 | 50 | 13.45 | Yes |
| 64 | 135 | 125 | 3500 | 1    | 20 | 50 | 15.08 | Yes |
| 65 | 135 | 175 | 3500 | 1.2  | 20 | 50 | 16.15 | Yes |
| 66 | 135 | 175 | 4000 | 1    | 16 | 75 | 16.32 | Yes |
| 67 | 140 | 150 | 3500 | 1.2  | 20 | 50 | 17.13 | Yes |
| 68 | 140 | 175 | 4000 | 1    | 20 | 50 | 15.56 | Yes |
| 69 | 140 | 175 | 4500 | 1    | 20 | 50 | 15.93 | Yes |
| 70 | 140 | 200 | 2500 | 1    | 20 | 25 | 8.15  | No  |
| 71 | 140 | 200 | 4000 | 0.8  | 20 | 50 | 15.95 | Yes |
| 72 | 145 | 125 | 3000 | 1.2  | 16 | 50 | 17.42 | Yes |
| 73 | 145 | 150 | 3500 | 0.8  | 16 | 50 | 15.48 | Yes |
| 74 | 145 | 150 | 4000 | 1    | 20 | 75 | 12.99 | Yes |
| 75 | 145 | 175 | 3000 | 1    | 16 | 25 | 16.73 | Yes |
| 76 | 145 | 175 | 4000 | 0.8  | 20 | 75 | 16.49 | Yes |
| 77 | 150 | 150 | 3000 | 1    | 20 | 50 | 7.75  | No  |
| 78 | 150 | 175 | 3500 | 1    | 16 | 50 | 17.66 | Yes |
| 79 | 155 | 150 | 3000 | 1.2  | 20 | 50 | 10.98 | No  |
| 80 | 144 | 166 | 3270 | 1.15 | 19 | 44 | 14.98 | Yes |
| 81 | 145 | 130 | 3090 | 1.1  | 17 | 29 | 17.41 | Yes |
| 82 | 150 | 141 | 3300 | 1.1  | 20 | 50 | 14.86 | Yes |
| 83 | 149 | 170 | 3160 | 1    | 20 | 35 | 14.43 | Yes |
| 84 | 142 | 126 | 3230 | 1.05 | 16 | 40 | 16.90 | Yes |
| 85 | 140 | 155 | 3200 | 1.2  | 17 | 33 | 13.57 | Yes |
| 86 | 149 | 148 | 3120 | 1    | 18 | 37 | 14.34 | Yes |
| 87 | 143 | 163 | 3340 | 1.15 | 19 | 31 | 17.32 | Yes |
| 88 | 148 | 137 | 3410 | 1.15 | 18 | 46 | 13.02 | No  |
| 89 | 145 | 145 | 3490 | 1.2  | 19 | 27 | 18.45 | Yes |
| 90 | 141 | 134 | 3050 | 1.2  | 20 | 48 | 15.28 | Yes |

| | | | | | | | |
|---|---|---|---|---|---|---|---|
| 91 | 146 | 159 | 3450 | 1.05 | 16 | 38 | 16.76 | Yes |
| 92 | 141 | 174 | 3010 | 1 | 16 | 42 | 11.53 | Yes |
| 93 | 147 | 152 | 3380 | 1.2 | 17 | 25 | 16.71 | Yes |
| 94 | 147 | 151 | 3280 | 1.15 | 17 | 39 | 14.03 | Yes |
| 95 | 140 | 150 | 3500 | 1.2 | 20 | 50 | 15.96 | Yes |
| 96 | 140 | 150 | 3500 | 1 | 20 | 50 | 10.74 | Yes |
| 97 | 145 | 125 | 3000 | 1.2 | 16 | 50 | 16.62 | Yes |
| 98 | 150 | 175 | 3500 | 1 | 16 | 50 | 14.99 | Yes |
| 99 | 145 | 125 | 3500 | 1 | 20 | 50 | 10.07 | Yes |

b. The process conditions *vs I-V* characteristics of top-performing devices

We documented the process conditions and the *I-V* characteristics in Table S5 for the top performers (PCE ≥17%). The histogram distribution of these 10 process conditions (binned into specific levels) are shown in Fig. S5. The corresponding *I-V* curves are shown in Fig. S6.

Table S5: The process conditions and the *I-V* characteristics for the 10 top-performer (PCE ⩾17% devices)

| **Process condition** | Temperature [°C] | Speed [cm/s] | Spray flowrate [mL/min] | Plasma height [cm] | Plasma gas flowrate [L/min] | Plasma duty cycle [%] | $j_{SC}$ [mA/cm²] | $V_{OC}$ [V] | FF [-] | **PCE [%]** |
|---|---|---|---|---|---|---|---|---|---|---|
| **11** | 145 | 125 | 3.5 | 1 | 25 | 50 | 23.9 | 1.047 | 0.68 | **17.0** |
| **43** | 135 | 150 | 3.5 | 1.2 | 20 | 25 | 23.7 | 0.919 | 0.78 | **17.0** |
| **44** | 135 | 150 | 4.0 | 0.8 | 20 | 50 | 22.4 | 1.007 | 0.77 | **17.4** |
| **46** | 135 | 175 | 4.0 | 1.2 | 20 | 50 | 22.2 | 0.997 | 0.77 | **17.0** |
| **67** | 140 | 150 | 3.5 | 1.2 | 20 | 50 | 22.7 | 0.980 | 0.77 | **17.1** |
| **72** | 145 | 125 | 3.0 | 1.2 | 16 | 50 | 23.0 | 0.985 | 0.77 | **17.4** |
| **78** | 150 | 175 | 3.5 | 1 | 16 | 50 | 22.7 | 1.023 | 0.76 | **17.7** |
| **81** | 145 | 130 | 3.1 | 1.1 | 17 | 29 | 23.2 | 0.975 | 0.77 | **17.4** |
| **87** | 143 | 163 | 3.3 | 1.15 | 19 | 31 | 21.7 | 1.000 | 0.80 | **17.3** |
| **89** | 145 | 145 | 3.5 | 1.2 | 19 | 27 | 23.6 | 1.004 | 0.78 | **18.5** |

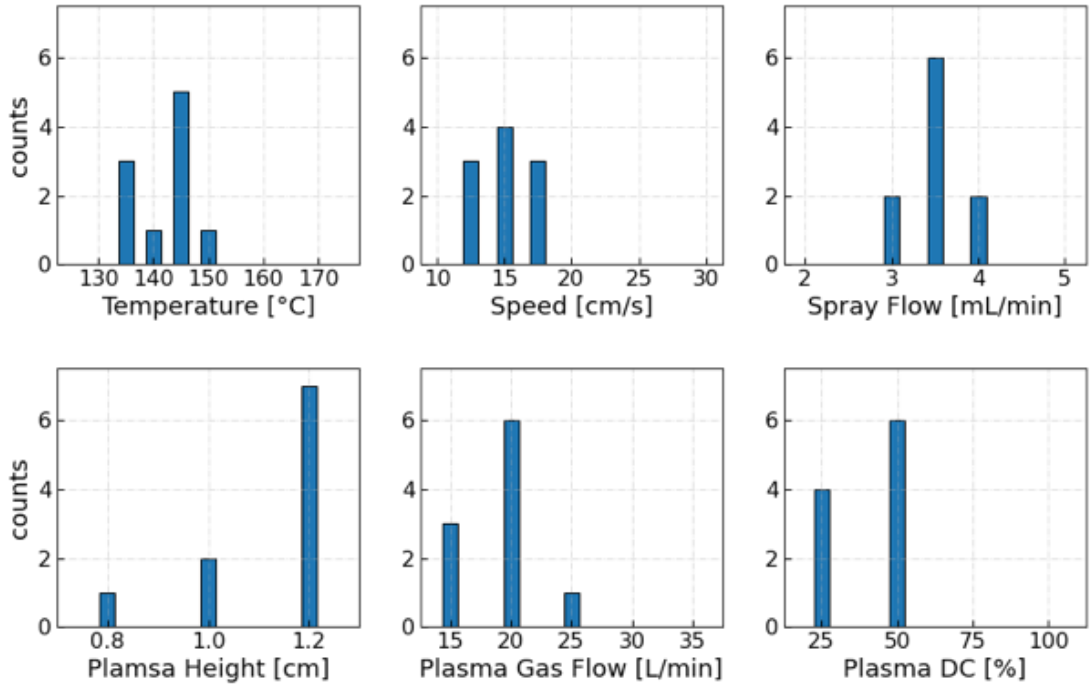

Figure S5: The histogram of the process conditions for the 10 top-performer (PCE ≥17% devices).

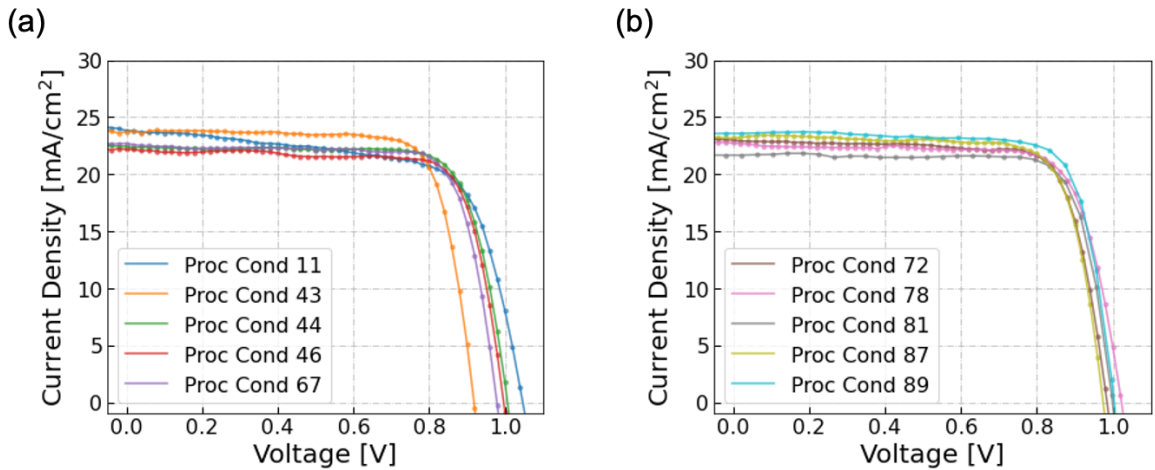

Figure S6: The in-house measured *I-V* curves of the top efficiency devices (≥17%) that were produced in this work. Process condition no. 89 achieves the champion efficiency of 18.5% efficiency.

Fig. S7 shows the spectral-dependent external quantum efficiency (EQE) of the device under process condition 89. We integrated the EQE curve with the AM1.5G solar spectrum and obtained a current density value of 23.0 mA/cm². The main reason for the discrepancy between the EQE-integrated current density and the short-circuit current density may possibly come from device degradation in the storage, insufficient light soaking prior to EQE measurement, slight spectral mismatch between AM1.5G and solar simulator and so on.

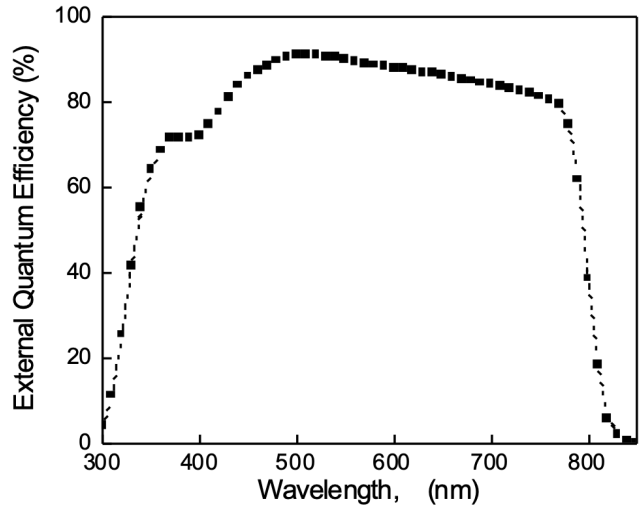

Figure S7: The external quantum efficiency (EQE) of the champion device fabricated under process condition 89. The integrated current density 23.02 mA/cm$^2$ (weighted under AM1.5G)

We selected several samples in the same batch as the best performing solar cells, and obtained the optical microscope images (under reflection mode) and the maps of light beam induced current. We found that non-uniformity is one of the key factors affecting both the LBIC and the optical images (see Figs. 8a & 8e and 8d & 8h). Please note that both thickness variations (which affect optical absorption) and carrier collection is reflected in the LBIC maps.

Table S6: The selected comparison of the process conditions and device *I-V* characteristic parameters in the last batch of experimental iteration

| ML Condition | Sample | Temperature [°C] | Speed [mm/s] | Spray flowrate [mL/min] | Plasma Height [cm] | Plasma Gas Flow [L/min] | Plasma DC [%] | $j_{SC}$ [mA/cm$^2$] | $V_{OC}$ [V] | *FF* [-] | PCE [%] |
|---|---|---|---|---|---|---|---|---|---|---|---|
| 84 | F14R | 142 | 126 | 3.23 | 1.05 | 16 | 40 | 24.1 | 1.062 | 0.66 | 16.9 |
| 81 | F19R | 145 | 130 | 3.09 | 1.1 | 17 | 29 | 23.2 | 0.975 | 0.77 | 17.4 |
| 89 | F22R | 145 | 145 | 3.49 | 1.2 | 19 | 27 | 23.6 | 1.004 | 0.78 | 18.5 |
| 91 | F28R | 146 | 159 | 3.45 | 1.05 | 16 | 38 | 22.3 | 0.991 | 0.76 | 16.8 |

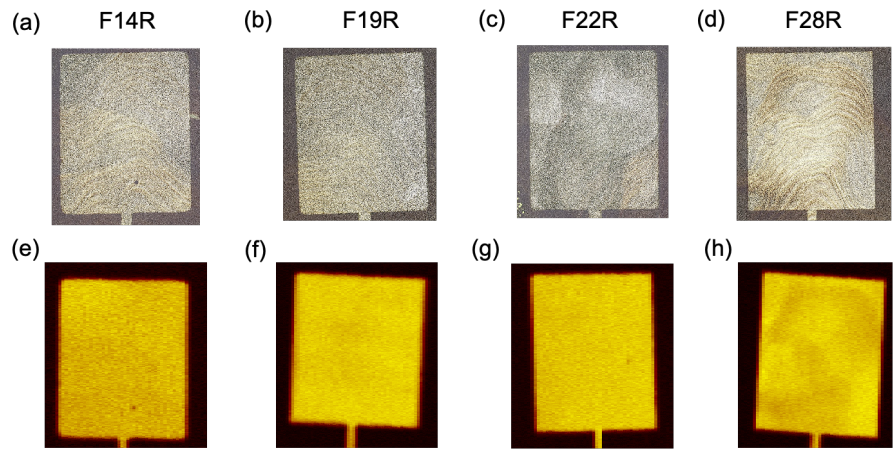

Figure S8: The optical images (in a, b, c, d) and the light-beam-induced-current maps (in e, f, g, h) are shown for four devices fabricated under different process conditions.

## 3. Compare Sequential Learning vs. Design of Experiment for Device Optimization

*3.1. Deep Dive to the Entire Optimization Space with Random Sampling and Full Factorial Sampling*

To understand the overall efficiency distribution of the entire parameter space, we ran both the random sampling of 100,000 process conditions and the full factorial sampling of 41,580 process conditions (see Table I in the main text). The predicted efficiency distributions are shown in Fig. S9. The full factorial sampling and the 100k random sampling have very similar distributions (with only minor difference for the range of 10 – 13%). Based on the 100k random samples, we estimated the top 5 percentile is around 0.8 at the normalized PCE (14.2%/17.7%) in the predicted PCE (marked by blue line in Fig. S9). This top one percentile mark is also used in Fig. 6 from the main text to compare different sampling methods. Again, note that the "teacher" model slightly underestimates the top performers. Therefore, we normalized the predicted efficiency by 17.7% in Figs. 6b and 6c (from the main text), which is the highest predicted efficiency among all the experimental conditions in this work.

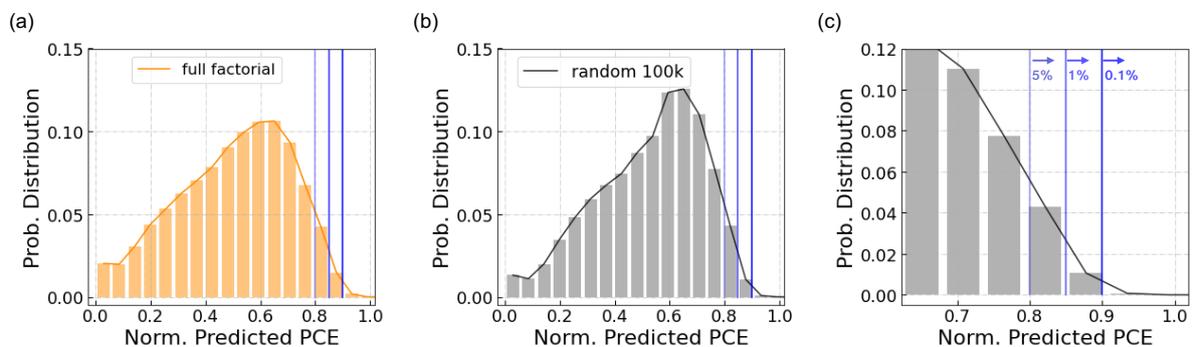

Figure S9: The histogram distribution of the predicted efficiencies by the "teacher" regression model. (a) The full factorial sampling of all the grid levels defined in Table I. (b) The randomly sampled 100k process conditions. (c) The zoom-in view of (b) to show the blue lines of the top 5%, 1% and 0.1% in the random 100k samples, which are approximately corresponding to 0.8, 0.85 and 0.9 respectively in the normalized PCE.

*3.2. Design of Experiment Method 1: Latin Hypercube Sampling (LHS)*

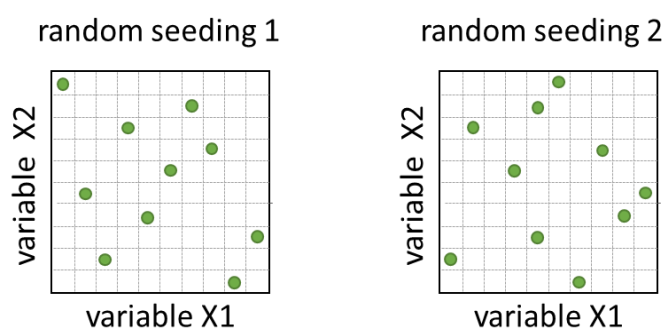

Figure S10: Schematic illustration of Latin hypercube sampling (LHS) with different runs of random seeding in an exemplary 2D space. The green solid circles are the sampled conditions.

Latin hypercube sampling is known as a low-discrepancy pseudo-random sampling method. The parameter space is stratified into a hypercube with *m* levels in every dimension, and samples are picked sequentially so that each parameter level is only sampled once. At every iteration except for the last sample, the parameters are chosen randomly in the unoccupied hypercubes. As shown in Fig. S10, the pseudo-randomness can result in different samples. When being run many times, the statistical distribution will converge to the same as the random sampling.

*3.3. Design of Experiment Method 2: One Variable at a Time Sampling (OVATS)*

The one-variable-at-a-time method is one of the most popular methods for the experimental planning in academic research. The idea is simply illustrated in Fig. S11. The initial round will sample all levels in one variable while the rest variables are set to randomly picked values, and the best condition can be found. Then, the first variable is fixed to that best condition while all levels in another variable are sampled. The sequence of the sampled variables is determined at random. This procedure iterates through all the variables, and it repeats until no improvement can be achieved. In that case, we subdivide the grid, and it restarts the sampling procedure (see Round 4 in Fig. S11).

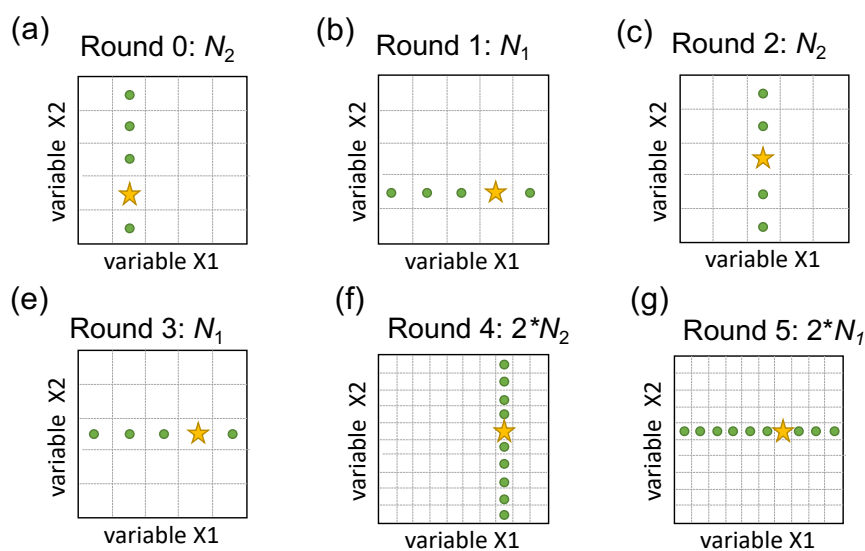

Figure S11: Schematic illustration of one variable at a time sampling (OVATS). The green solid circles are sampled conditions in that round. The yellow star is the best condition of all the samples until that specific round.

*3.4. Design of Experiment Method 3: Factorial Sampling in Progressive Grid Division (FS-PGS)*

The full factorial sampling suffers from the curse of dimensionality, which means the number of samples becomes infeasible to achieve experimentally as the dimensionality increases. To sample $m$ levels in $n$-variable optimization, we need to sample $m^n$ conditions in a full grid. For example, 5 levels in 6-variable optimization result in 15625 conditions. In practice, we can use two-level grid sampling sequentially to achieve a more efficient design of the experiments. After each round, we rely on the location of the best device in the parameter space to define the new optimization grid in the next round.

An initial sampling (or Round 0) is started with a full two-level grid in each variable, *i.e.*, high (H) and low (L) conditions. The total number of samples in Round 0 is $2^n$, where $n$ is the total number of variables in the optimization (*e.g.*, six variables in this work). After Round 0 experiments, we see the best condition is found at the (L, H) condition. Therefore, the new reduced space of Round 1 is moved toward the best condition in Round 0. The parameter space keeps reducing after each round until the minimum sampling space is reached (see Round3 in Fig. S12e). After that, the sampling is repeated iteratively with the minimum sampling space, until the location of the best device is not changing (see Round 4 in Fig. S12f). In this case, the sampling space in the next round is increased to explore a broader region (see Round 5 in Fig. S12g). This is just one way to implement the factorial sampling with progressive grid subdivision, but the key concept is to exploit the regions where the best device is found.

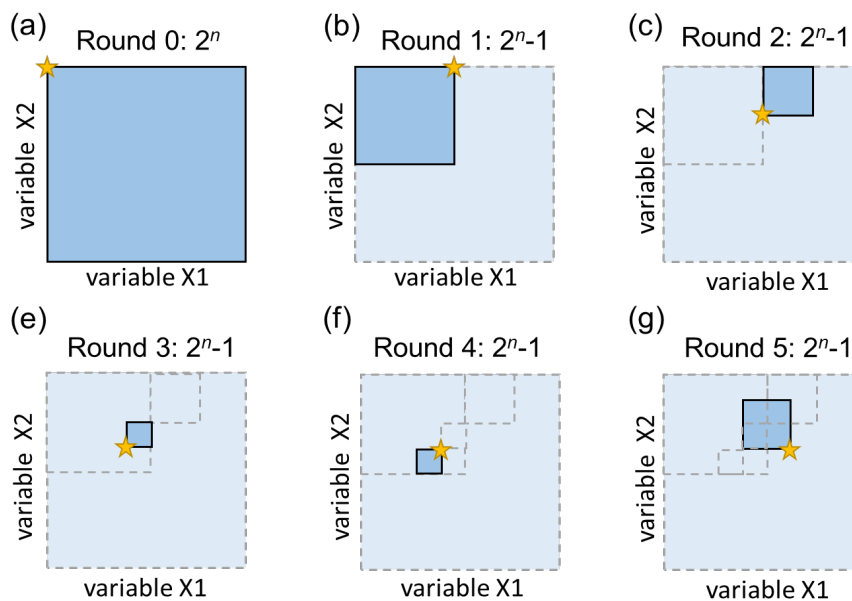

Figure S12: Schematic illustration of an exemplary implementation of the factorial sampling with progressive grid subdivision. The darker blue box (with black lines) is the parameter space for two-level factorial sampling in the specific round. The dashed gray boxes are the sampled space in previous rounds, where corner points or intersections are the conditions that have already been sampled. The yellow star is the best device of all the samples. The smallest box is 1/8 of the entire parameter space, so the total level of each variable is nine. In this illustration, the total number of sampling points is 379 after all six rounds.

*3.5 Benchmarking: the BO Methods vs Model-Free Design of Experiment Methods*

In addition to the Fig. 6 in the main text, we plotted the several virtual benchmarking results in full details with separate figures.

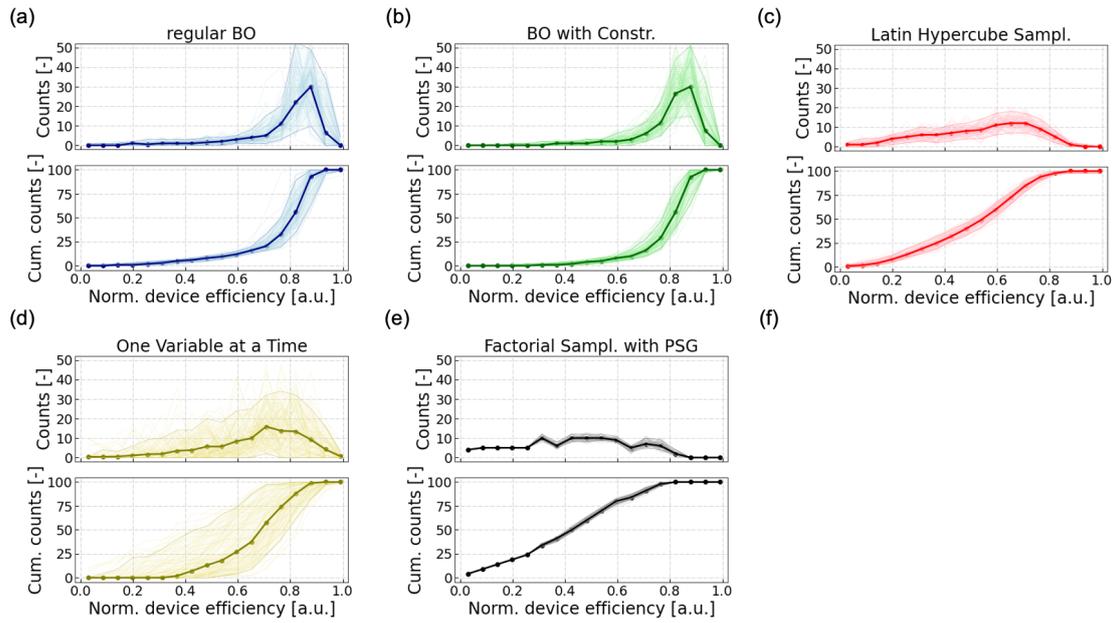

Figure S13: The histogram distribution within the first 100 experimental budget by the simulated process optimization. The lighter lines in each figure represents the 100 separate runs with random seeding. The darker solid line represents the median of the 100 runs, and the dashed lines indicates the confidence interval of 5% – 95%.

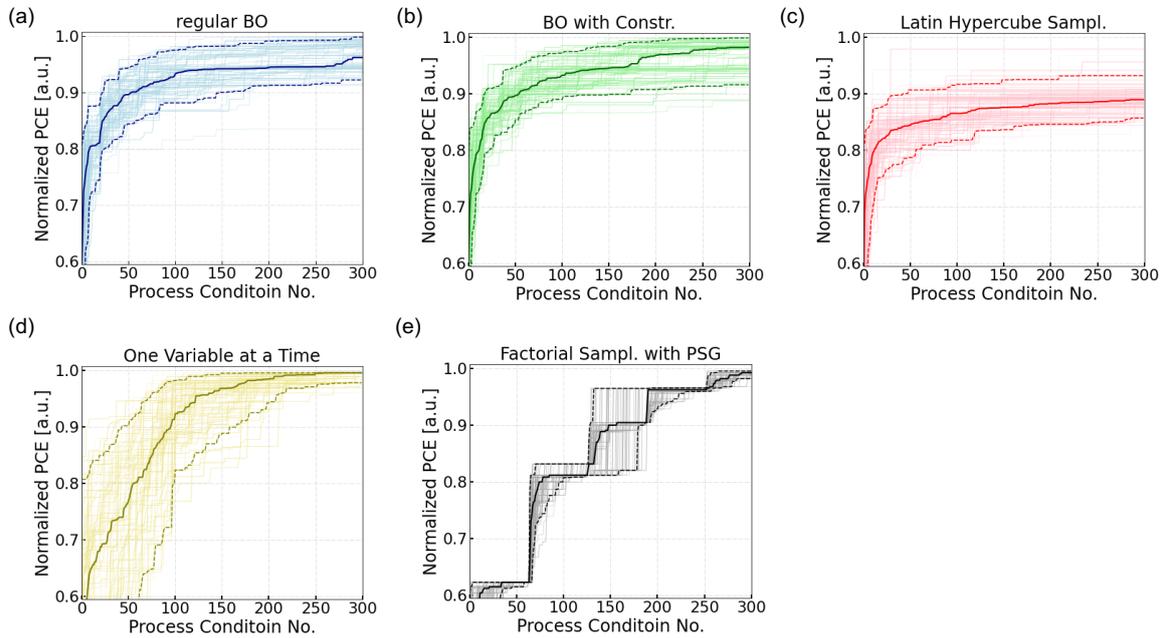

Figure S14: The best efficiency at a given number of process conditions by the simulated process optimization. The lighter lines in each figure represents the 100 separate runs with random seeding. The darker solid line represents the median of the 100 runs, and the dashed lines indicates the confidence interval of 5% – 95%.

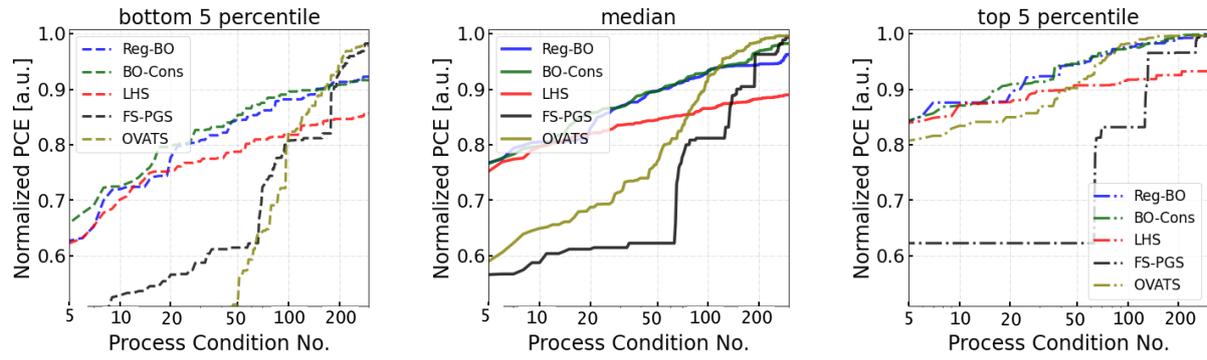

Figure S15: The comparison of the best PCE vs the number of process conditions explored among the different sampling methods. Different percentiles are plotted in the (a), (b) and (c) respectively.

References for SI:

[1] Niranjan Srinivas, Andreas Krause, Sham M. Kakade, and Matthias W. Seeger, "Information-Theoretic Regret Bounds for Gaussian Process Optimization in the Bandit Setting," *IEEE Transactions on Information Theory*, vol. 58, no. 5, pp. 3250–3265, May 2012.

[2] J. Mockus, V. Tiesis, and A. Zilinskas, "The application of Bayesian methods for seeking the extremum," in *Towards Global Optimisation*, vol. 2, L.C.W.Dixon and G.P. Szego, Eds. North-Holand, 1978.

[3] B. Rohr, H. S. Stein, D. Guevarra, Y. Wang, J. A. Haber, M. Aykol, S. K. Suram, and J. M. Gregoire, "Benchmarking the acceleration of materials discovery by sequential learning," *Chemical Science*, vol. 11, no. 10, pp. 2696–2706, Mar. 2020.

[4] N. Rolston, W. J. Scheideler, A. C. Flick, J. P. Chen, H. Elmaraghi, A. Sleugh, O. Zhao, M. Woodhouse, and R. H. Dauskardt, "Rapid Open-Air Fabrication of Perovskite Solar Modules," *Joule*, vol. 4, no. 12, pp. 2675–2692, Dec. 2020.